\documentclass{article}

\usepackage{arxiv}

\usepackage[utf8]{inputenc} 
\usepackage[T1]{fontenc}    
\usepackage{hyperref}       
\usepackage{url}            
\usepackage{booktabs}       
\usepackage{amsfonts}       
\usepackage{amsmath}        
\usepackage{nicefrac}       
\usepackage{microtype}      
\usepackage{lipsum}		
\usepackage{graphicx}
\usepackage{natbib}
\usepackage{doi}
\usepackage{algorithm}
\usepackage{algpseudocode}
\usepackage{longtable}
\usepackage{float} 
\usepackage{tikz}
\usepackage{rotating}
\usepackage{afterpage}

\title{An Interpretable Ensemble-Based Generative Framework for Anomaly Detection in High-Dimensional Financial Time Series}

\date{} 					

\author{ \href{https://orcid.org/0000-0002-6634-9016}{\includegraphics[scale=0.06]{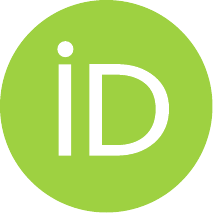}\hspace{1mm}Waldyn Martinez} \\
	Department of Information Systems \& Analytics\\
    Farmer School of Business \\
	Miami University\\
	Oxford, OH 45056 \\
	\texttt{martinwg@miamioh.edu} \\
	  \\
}

\renewcommand{\shorttitle}{An Interpretable Ensemble-Based Generative Framework for Anomaly Detection}

\hypersetup{
pdftitle={A template for the arxiv style},
pdfsubject={q-bio.NC, q-bio.QM},
pdfauthor={David S.~Hippocampus, Elias D.~Striatum},
pdfkeywords={First keyword, Second keyword, More},
}

\begin{document}
\maketitle

\begin{abstract}
Detecting structural instability and anomalies in high-dimensional financial time series is challenging due to complex temporal dependence and evolving cross-sectional structure. We propose ReGEN-TAD, an interpretable generative framework that integrates modern machine learning with econometric diagnostics for anomaly detection. The model combines joint forecasting and reconstruction within a refined convolutional--transformer architecture and aggregates complementary signals capturing predictive inconsistency, reconstruction degradation, latent distortion, and volatility shifts. Robust calibration yields a unified anomaly score without labeled data. Experiments on synthetic and financial panels demonstrate improved robustness to structured deviations while enabling economically coherent factor-level attribution.
\end{abstract}


\section{Introduction}
\label{sec:introduction}

Anomaly detection in high-dimensional multivariate time series is a central problem in modern financial econometrics, where large cross-sectional panels of asset returns exhibit complex dependence structures, nonstationary dynamics, and regime shifts. Observations arrive as temporally ordered vectors whose components interact through evolving correlation patterns, volatility clustering, and latent factor structure. Anomalies may appear as isolated spikes, sustained deviations, gradual structural change, or contextual departures relative to an estimated baseline regime, to name a few. The combination of temporal dependence, high dimensionality, and structural instability makes reliable detection statistically demanding and economically consequential, particularly when the objective is not only identification but also structural interpretation.

Classical anomaly detection techniques, including distance-based methods, density estimators, and isolation-based approaches, typically deteriorate in high-dimensional panel settings due to the curse of dimensionality and the breakdown of independence assumptions. Recent advances in machine learning have introduced powerful sequence models, such as recurrent autoencoders, variational architectures, and attention-based transformers, that flexibly capture nonlinear temporal structure. Yet most existing approaches rely on a single diagnostic signal, commonly reconstruction error or predictive likelihood, implicitly assuming that one dimension of model failure sufficiently characterizes abnormality. In financial panels where anomalies may affect subsets of assets, alter cross-sectional dependence, or unfold gradually across horizons, reliance on a single criterion can be unstable and economically opaque.

Generative modeling provides a flexible framework for learning joint temporal and cross-sectional structure in multivariate financial data. Architectures combining convolutional, recurrent, and attention mechanisms are capable of representing both local dynamics and long-range interactions. However, detection based on a single generative score remains sensitive to misspecification, estimation error, and contamination in training samples. Subtle but persistent deviations may be absorbed into the learned representation of normality, reducing sensitivity to structural change. Moreover, most generative frameworks do not explicitly distinguish predictive inconsistency, reconstruction degradation, and latent representation distortion as separate indicators of instability, limiting their usefulness for economic interpretation and policy-relevant diagnostics.

To address these limitations, we introduce ReGEN-TAD (\textit{Refined Generative Ensemble for Temporal Anomaly Detection}), a unified framework that integrates modern generative machine learning with econometric diagnostics for structural instability in high-dimensional panels. The core contribution lies in jointly modeling short-horizon predictive consistency and reconstruction fidelity within a modular ensemble architecture that aggregates complementary signals of deviation. Rather than relying on a single score, the method combines prediction errors, reconstruction loss, latent-space geometry, temporal instability, and volatility variation, each capturing a distinct manifestation of abnormal behavior. This ensemble design enhances robustness across heterogeneous anomaly mechanisms while preserving interpretability consistent with economic structure.

The proposed framework contributes along three dimensions. First, it formalizes predictive degradation across horizons as a measurable indicator of structural instability, linking forecast inconsistency to economically meaningful regime change. Second, it embeds ensemble aggregation within a shared generative representation, mitigating masking effects caused by contamination and nonstationarity. Third, it incorporates structured factor-level attribution directly into the detection mechanism, enabling recovery of economically coherent cross-sectional perturbations without imposing additional post hoc explanation layers. Together, these elements illustrate how modern machine learning architectures can be integrated with econometric reasoning to balance predictive flexibility and structural interpretability.

At the core of ReGEN-TAD is a generative backbone that simultaneously supports short-horizon forecasting and input reconstruction. A convolutional front-end feeds into a transformer-based encoder and recurrent temporal pathway, producing compact latent representations of rolling windows. A two-stage refinement mechanism adjusts initial forecasts using residual information conditioned on the latent state, improving predictive stability and sharpening the residual structure. This refinement increases sensitivity to gradual distributional drift and evolving dependence patterns that often characterize financial regime change.

Anomaly detection proceeds by constructing multiple diagnostic scores derived from refined forecasts and latent representations. These scores are normalized using robust, scale-free statistics and aggregated through an adaptive calibration procedure that estimates baseline behavior directly from the data. The procedure remains fully unsupervised and accommodates moderate contamination in the baseline sample. By explicitly combining predictive and reconstructive diagnostics, the framework operationalizes instability monitoring across multiple dimensions of deviation in high-dimensional financial panels.

Beyond detection, ReGEN-TAD provides structured factor-level attribution. For each anomalous window, the aggregate anomaly score is decomposed into variable-specific contributions derived from standardized deviations relative to the estimated baseline regime and model-implied sensitivities from the reconstruction pathway. This mechanism links detected instability to economically coherent subsets of assets, enabling recovery of sector-level or factor-level perturbations without imposing additional explainability assumptions. Attribution is embedded within the detection architecture, ensuring consistency between identification and interpretation.

The framework is evaluated through extensive simulation studies and benchmark datasets spanning diverse anomaly mechanisms, including mean shifts, trend changes, volatility spikes, contextual deviations, and jump disturbances across multiple dimensional regimes. Results show that ReGEN-TAD improves robustness to structured and gradual deviations while maintaining competitive performance for isolated anomalies. Structured attribution experiments further demonstrate recovery of economically meaningful cross-sectional perturbations under controlled injections, reinforcing the view that generative machine learning methods can be deployed in a manner consistent with econometric interpretability.

The remainder of the paper is organized as follows. Section~\ref{sec:preliminaries} reviews related literature. Section~\ref{sec:method} presents the ReGEN-TAD framework, including the generative architecture, refinement mechanism, ensemble scoring, calibration strategy, and attribution methodology. Section~\ref{sec:experiments} reports empirical results, and Section~\ref{sec:conclusion} concludes.

\section{Preliminaries and Related Work}
\label{sec:preliminaries}

This section reviews the methodological foundations underlying anomaly detection in high-dimensional financial time series. We first summarize structural characteristics of financial data that complicate detection, then discuss econometric and machine-learning approaches for identifying abnormal behavior. Finally, we review advances in deep generative modeling that motivate the proposed ensemble-driven framework. Throughout, the emphasis is on instability detection in cross-sectional panels under limited structural assumptions.

\subsection{Characteristics of Financial Time Series}
\label{subsec:financial_characteristics}

Financial time series exhibit structural properties that fundamentally complicate anomaly detection. Returns and related variables are serially dependent, cross-sectionally correlated, heavy-tailed, and heteroskedastic \citep{cont2001empirical}. These features violate classical assumptions of independence and homoskedastic Gaussian noise, particularly in high-dimensional panels where dependence propagates across assets, sectors, and latent factors. From an econometric perspective, anomaly detection in such environments is closely related to identifying instability in conditional means, variances, and dependence structures.

Trending and low-frequency components further complicate inference. As emphasized by \citet{phillips2001trending,phillips2005challenges}, trending behavior is empirically pervasive yet difficult to characterize probabilistically. Even visually apparent trends often lack a clear structural interpretation, and misspecification can distort inference and forecasting. This ambiguity motivates detection procedures that remain robust to unknown trend structures and evolving baselines rather than relying on tightly parameterized models.

Anomalies arise in multiple forms reflecting both transient and structural phenomena. Isolated extreme observations may correspond to liquidity shocks or information arrivals, whereas persistent deviations often signal regime changes, volatility clustering, or systemic stress. In multivariate settings, anomalies frequently emerge collectively through coordinated movements across subsets of assets rather than through marginal outliers. Such behavior challenges detectors based on univariate statistics or fixed marginal thresholds.

Classical econometric models, including ARMA and GARCH-type specifications, model conditional means and variances under relatively stable regimes \citep{bollerslev1986generalized}. While valuable for forecasting and inference, they are not designed to detect atypical behavior under evolving structural conditions. Regime-switching models partially address this limitation but typically impose strong assumptions on the number and nature of latent states \citep{hamilton1989regime}. More generally, forecast performance deteriorates sharply under structural breaks and evolving dynamics \citep{clements1998forecasting,clements2006forecasting}, suggesting that predictive degradation itself may serve as an instability diagnostic.

Evidence from financial forecasting further underscores the fragility of single-model approaches. Using a diverse set of statistical and machine-learning models, \citet{dingli2017financial} document substantial sensitivity of predictive performance to aggregation level, regime, and horizon. No model performs uniformly well across assets or time scales. These findings imply that deviations from a single forecast residual or likelihood measure cannot reliably characterize abnormality in unstable financial environments.

\subsection{Anomaly Detection in Financial and Econometric Settings}

Anomaly detection seeks to identify observations or temporal segments that deviate from typical system behavior \citep{chandola2009anomaly}. In financial econometrics, related problems appear as outlier detection, jump detection, structural break analysis, and systemic risk identification. Classical approaches include changepoint tests \citep{bai1998estimating}, jump detection procedures \citep{barndorff2004power}, and volatility-based diagnostics that flag unusually large innovations.

Nonparametric and machine-learning approaches relax assumptions and often scale more effectively with dimensionality. Distance-based methods such as $k$-nearest neighbors identify observations in sparse regions of feature space \citep{ramaswamy2000efficient}. Density-based methods, including LOF, compare local densities to detect deviations \citep{breunig2000lof}. Ensemble strategies such as Isolation Forest exploit random partitioning to isolate rare observations efficiently \citep{liu2008isolation}. Distribution-free procedures, including ECOD, enhance robustness under heavy tails and heterogeneous distributions \citep{li2022ecod}. Although originally designed for static data, these methods are often adapted to time-series settings through windowing or residual extraction. A comprehensive overview is provided by \citet{schmidl2022anomaly}.

Dimension-reduction techniques have also been employed in high-dimensional financial settings. Principal component analysis (PCA) detects anomalies through deviations in low-variance components or reconstruction errors in reduced subspaces \citep{crepey2022anomaly}. While effective for linear dependence structures, such methods may fail to capture nonlinear dynamics, evolving cross-sectional relationships, or collective deviations driven by latent structural change.

Despite their flexibility, many existing detectors treat observations as conditionally independent or rely on static feature representations. In time-series contexts, ignoring sequential dependence can produce excessive false positives or miss contextual anomalies. Moreover, methods based on a single diagnostic principle often perform well for specific anomaly types but deteriorate under regime shifts. These limitations motivate ensemble strategies that integrate multiple instability diagnostics to improve robustness in nonstationary environments \citep{aggarwal2017outlier}.

\subsection{Deep Learning and Representation-Based Approaches}

Deep learning provides flexible tools for modeling nonlinear temporal and cross-sectional dependence. Recurrent networks, temporal convolutional architectures, and attention-based transformers \citep{vaswani2017attention} have demonstrated strong forecasting and representation capabilities \citep{lim2021temporal, zhou2021informer, zeng2023financial}. Hybrid convolution-transformer designs capture localized patterns and long-range dependencies within unified architectures. However, most such models are optimized for supervised prediction rather than unsupervised instability detection.

In anomaly detection, deep models typically learn representations of normal behavior and flag deviations through reconstruction errors, predictive residuals, or likelihood discrepancies. This framework implicitly assumes predominantly clean training data. In financial time series, where regime changes and structural breaks are common, contaminated training samples may lead the model to internalize abnormal dynamics, weakening subsequent discrimination.

Deep anomaly detection methods can be broadly categorized as reconstruction-based, prediction-based, probabilistic, or hybrid approaches \citep{schmidl2022anomaly,chalapathy2019deep}. In financial applications, the literature remains relatively limited and often focuses on fraud or transactional data framed as supervised classification problems \citep{phua2010comprehensive,paula2016deep}. Applications to unsupervised detection in high-dimensional asset panels are comparatively sparse.

Reconstruction-based models, particularly autoencoders and variational autoencoders \citep{kingma2013auto}, remain widely used due to their intuitive premise: typical behavior can be reconstructed accurately, whereas anomalies generate larger errors. While expressive, such models share a common limitation: persistent or gradual structural deviations may be partially absorbed into the learned representation of normality. This masking effect is especially pronounced when anomalies reflect evolving dependence structure rather than isolated point outliers.

\subsection{Motivation for an Ensemble-Driven Generative Framework}

The preceding discussion highlights a structural gap in the anomaly-detection literature for financial panels. Existing approaches typically operationalize abnormality through a single diagnostic principle: forecast residuals, reconstruction error, density deviation, or variance instability. While each captures a particular channel of structural change, none is sufficient in isolation when conditional means, volatility dynamics, and cross-sectional dependence evolve simultaneously.

From an econometric perspective, anomaly detection in high-dimensional financial systems is fundamentally an instability-monitoring problem. Structural change may manifest through predictive breakdown, dispersion shifts, latent factor reconfiguration, or coordinated cross-asset movements. Because these mechanisms operate along different statistical dimensions, reliance on a single residual or likelihood-based statistic can lead to both masking and spurious detection under regime shifts.

This observation motivates a framework that treats anomaly detection as joint monitoring across complementary structural channels. Rather than committing to a single notion of deviation, we construct multiple diagnostics derived from a shared representation of temporal and cross-sectional dynamics. A generative backbone provides this representation by learning a latent embedding of the system's evolving state. Distinct instability measures can then be formulated as functionals of that embedding, each targeting a different aspect of structural departure.

The resulting ensemble is flexible by design: new diagnostics can be incorporated or existing ones removed without altering the underlying architecture. Detection becomes a calibrated aggregation of complementary instability signals rather than the output of a single criterion. In this sense, the contribution is conceptual as well as methodological: anomaly detection is reframed as multi-dimensional structural monitoring embedded within a learned generative representation. Because each diagnostic is explicitly constructed and aggregated, the framework also supports component-level interpretability, including attribution of detected instability to specific factors or cross-sectional drivers. This perspective bridges econometric instability diagnostics with modern representation learning while preserving transparency at the level of underlying structural channels.

\section{Methodology}
\label{sec:method}

This section presents the proposed ReGEN-TAD framework. The methodology is designed for unsupervised anomaly detection in high-dimensional multivariate time series and explicitly addresses the challenges of nonstationarity, training contamination, and heterogeneous anomaly mechanisms discussed in Section~\ref{sec:preliminaries}. The framework consists of four main components: (i) a generative forecasting--reconstruction backbone with attention-based representation learning, (ii) a data purification stage for robust baseline estimation, (iii) an ensemble anomaly scoring mechanism that integrates multiple complementary diagnostics, and (iv) a decision layer based on rank- or threshold-based anomaly selection. An interpretable factor attribution procedure is provided for post-detection analysis.

\subsection{Problem Setup}

\begin{figure}[ht]
\centering
\resizebox{0.6\columnwidth}{!}{
\begin{tikzpicture}

\def\LX{5}
\def\PX{3}
\def\HX{3}

\draw (0,0) rectangle (\LX,\PX);

\node[above=0.3cm] at (\LX/2,\PX)
{$\mathbf{X}_t \in \mathbb{R}^{L \times p}$};

\draw (0,1) -- (\LX,1);
\draw (0,2) -- (\LX,2);

\draw (1,0) -- (1,\PX);
\draw (2,0) -- (2,\PX);
\draw (3,0) -- (3,\PX);
\draw (4,0) -- (4,\PX);

\node at (0.5,2.5) {$x_{t-L+1}^{(1)}$};
\node at (2.5,2.5) {$\cdots$};
\node at (4.5,2.5) {$x_{t}^{(1)}$};

\node at (0.5,1.5) {$\vdots$};
\node at (2.5,1.5) {$\ddots$};
\node at (4.5,1.5) {$\vdots$};

\node at (0.5,0.5) {$x_{t-L+1}^{(p)}$};
\node at (2.5,0.5) {$\cdots$};
\node at (4.5,0.5) {$x_{t}^{(p)}$};

\draw (\LX+2,0) rectangle (\LX+2+\HX,\PX);

\node[above=0.3cm] at (\LX+2+\HX/2,\PX)
{$\mathbf{F}_t \in \mathbb{R}^{H \times p}$};

\draw (\LX+2,1) -- (\LX+2+\HX,1);
\draw (\LX+2,2) -- (\LX+2+\HX,2);

\draw (\LX+3,0) -- (\LX+3,\PX);
\draw (\LX+4,0) -- (\LX+4,\PX);

\node at (\LX+2.5,2.5) {$x_{t+1}^{(1)}$};
\node at (\LX+3.5,2.5) {$\cdots$};
\node at (\LX+4.5,2.5) {$x_{t+H}^{(1)}$};

\node at (\LX+2.5,1.5) {$\vdots$};
\node at (\LX+3.5,1.5) {$\ddots$};
\node at (\LX+4.5,1.5) {$\vdots$};

\node at (\LX+2.5,0.5) {$x_{t+1}^{(p)}$};
\node at (\LX+3.5,0.5) {$\cdots$};
\node at (\LX+4.5,0.5) {$x_{t+H}^{(p)}$};

\draw[->, thick] (\LX,1.5) -- (\LX+2,1.5);

\end{tikzpicture}
}
\caption{
Rolling-window construction. The model ingests past $L$ observations
$\mathbf{X}_t$ and jointly predicts the future block $\mathbf{F}_t$ over horizon $H$.
}
\label{fig:rolling_setup}
\end{figure}
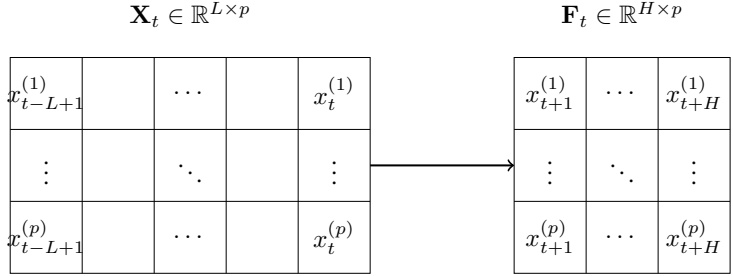

Let $\{\mathbf{x}_t\}_{t=1}^T$, with
$\mathbf{x}_t \in \mathbb{R}^p$, denote a $p$-dimensional
multivariate time series. Each vector
$\mathbf{x}_t = (x_t^{(1)}, \ldots, x_t^{(p)})^\top$
collects the $p$ cross-sectional components observed at time $t$. For each index $t$, we construct a rolling input window of length $L$,
\[
\mathbf{X}_t
=
(\mathbf{x}_{t-L+1}, \ldots, \mathbf{x}_t)
\in \mathbb{R}^{L \times p},
\]
together with the subsequent forecast block of length $H$,
\[
\mathbf{F}_t
=
(\mathbf{x}_{t+1}, \ldots, \mathbf{x}_{t+H})
\in \mathbb{R}^{H \times p}.
\]
As illustrated in Figure~\ref{fig:rolling_setup}, $\mathbf{X}_t$ contains the most recent $L$ observations up to time $t$, while $\mathbf{F}_t$ represents the future trajectory over the horizon $H$. The arrow emphasizes the temporal ordering: past information in $\mathbf{X}_t$ is used to infer the joint behavior of the subsequent block $\mathbf{F}_t$. The model is trained in a self-supervised manner. Given $\mathbf{X}_t$, it jointly
(i) reconstructs the input window and
(ii) predicts the future block $\mathbf{F}_t$.
No anomaly labels are used during training.

For each rolling window, an anomaly score $S_t$ is computed as an aggregation of multiple instability diagnostics derived from the joint behavior of $\mathbf{X}_t$ and $\mathbf{F}_t$. These diagnostics include forecasting discrepancies, reconstruction degradation, latent representation deviations, and temporal instability measures. The objective is to identify time indices $t$ at which the conditional dynamics linking $\mathbf{X}_t$ and $\mathbf{F}_t$ depart significantly from a learned baseline of typical system behavior.

\subsection{Generative Backbone Architecture}

The core of ReGEN-TAD is a generative backbone that jointly performs forecasting and reconstruction. The architecture combines convolutional feature extraction, attention-based temporal encoding, and recurrent modeling to capture both short-range and long-range dependencies. Each input window $\mathbf{X}_t$ is first normalized and passed through a stack of temporal convolutional layers to extract localized dependence patterns. Formally, the convolutional block applies $d$ filters of fixed temporal width along the time dimension of $\mathbf{X}_t$, operating jointly across all $p$ cross-sectional variables. 

\begin{figure}[H]
\centering
\begin{tikzpicture}

\def\LX{6}
\def\PX{3}

\draw (0,0) rectangle (\LX,\PX);

\node[above=0.3cm] at (\LX/2,\PX)
{$\mathbf{X}_t \in \mathbb{R}^{L \times p}$};

\draw (0,1) -- (\LX,1);
\draw (0,2) -- (\LX,2);

\foreach \i in {1,...,5}
    \draw (\i,0) -- (\i,\PX);

\fill[gray!25] (0,0) rectangle (3,\PX);

\foreach \i in {0,...,3}
    \draw (\i,0) -- (\i,\PX);
\draw (0,1) -- (3,1);
\draw (0,2) -- (3,2);

\node at (0.5,2.5) {$x_{t-L+1}^{(1)}$};
\node at (2.5,2.5) {$\cdots$};
\node at (5.5,2.5) {$x_{t}^{(1)}$};

\node at (0.5,1.5) {$\vdots$};
\node at (2.5,1.5) {$\ddots$};
\node at (5.5,1.5) {$\vdots$};

\node at (0.5,0.5) {$x_{t-L+1}^{(p)}$};
\node at (2.5,0.5) {$\cdots$};
\node at (5.5,0.5) {$x_{t}^{(p)}$};

\node[below=0.4cm] at (1.5,0)
{\small Temporal convolution (width $=3$)};

\draw[->, thick] (1.5,-0.8) -- (1.5,-1.8);

\draw (0.5,-2.2) rectangle (2.5,-3.2);
\node at (1.5,-2.7)
{$\mathbf{h}_{t-L+3} \in \mathbb{R}^{d}$};

\node at (1.5,-3.9)
{\small Feature vector at temporal location $\tau$};

\end{tikzpicture}
\caption{
\textbf{Temporal Convolution over the Input Window.}
A convolutional filter of width three is applied to the first three
time steps of $\mathbf{X}_t$, operating jointly across all $p$ features.
The shaded region represents the receptive field used to produce
a feature vector $\mathbf{h}_{t-L+3} \in \mathbb{R}^{d}$.
}
\label{fig:cnn_block}
\end{figure}
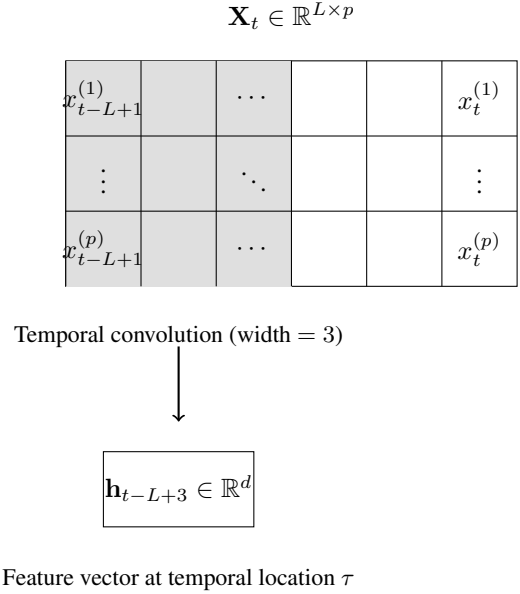

Figure~\ref{fig:cnn_block} provides an illustrative example of the temporal convolution operation, in which a filter of width three processes a contiguous block of time steps jointly across all $p$ features. Here, the shaded receptive field covering $\{\mathbf{x}_{t-L+1}, \mathbf{x}_{t-L+2}, \mathbf{x}_{t-L+3}\}$ yields the feature vector $\mathbf{h}_{t-L+3} \in \mathbb{R}^d$. More generally, each temporal location $\tau$ in the window is associated with a feature vector
\[
\mathbf{h}_{\tau} \in \mathbb{R}^d,
\]
constructed from a local neighborhood of observations preceding and including $\tau$. Stacking these vectors across the window produces a feature sequence
\[
\mathbf{H}_t = (\mathbf{h}_{t-L+1}, \ldots, \mathbf{h}_{t}) \in \mathbb{R}^{L \times d}.
\]
The convolutional representation maps the original $p$-dimensional observations into a learned feature space of dimension $d$ by applying shared temporal filters across local neighborhoods. This operation extracts short-run dynamic structure and nonlinear cross-sectional interactions while imposing parameter sharing that enhances statistical efficiency. The resulting features provide a compact summary of localized dependence patterns before longer-range dependencies are modeled by the subsequent attention and recurrent layers.

\noindent To enable global dependency modeling through self-attention, the convolutional features are projected into a $d$-dimensional embedding space and augmented with deterministic sinusoidal positional encodings. Because self-attention mechanisms are permutation-invariant, explicit temporal information must be incorporated to preserve the sequential structure characteristic of financial time series. 

\begin{figure}[H]
\centering
\begin{tikzpicture}

\def\W{3}
\def\H{3}

\draw (0,0) rectangle (\W,\H);

\node[above=0.3cm] at (1.5,\H)
{$\mathbf{H}_t \in \mathbb{R}^{L \times d}$};

\draw (0,1) -- (\W,1);
\draw (0,2) -- (\W,2);
\draw (1,0) -- (1,\H);
\draw (2,0) -- (2,\H);

\node at (0.5,2.5) {$h_{t-L+1}^{(1)}$};
\node at (1.5,2.5) {$\cdots$};
\node at (2.5,2.5) {$h_{t}^{(1)}$};

\node at (0.5,1.5) {$\vdots$};
\node at (1.5,1.5) {$\ddots$};
\node at (2.5,1.5) {$\vdots$};

\node at (0.5,0.5) {$h_{t-L+1}^{(d)}$};
\node at (1.5,0.5) {$\cdots$};
\node at (2.5,0.5) {$h_{t}^{(d)}$};

\node at (4,1.5) {$+$};

\draw (5,0) rectangle (5+\W,\H);

\node[above=0.3cm] at (6.5,\H)
{$\mathbf{P} \in \mathbb{R}^{L \times d}$};

\draw (5,1) -- (5+\W,1);
\draw (5,2) -- (5+\W,2);
\draw (6,0) -- (6,\H);
\draw (7,0) -- (7,\H);

\node at (5.5,2.5) {$p_{1}^{(1)}$};
\node at (6.5,2.5) {$\cdots$};
\node at (7.5,2.5) {$p_{L}^{(1)}$};

\node at (5.5,1.5) {$\vdots$};
\node at (6.5,1.5) {$\ddots$};
\node at (7.5,1.5) {$\vdots$};

\node at (5.5,0.5) {$p_{1}^{(d)}$};
\node at (6.5,0.5) {$\cdots$};
\node at (7.5,0.5) {$p_{L}^{(d)}$};

\draw[->, thick] (4,1) -- (4,0);

\draw (2.5,-4) rectangle (2.5+\W,\H-4);

\node[above=0.3cm] at (4,-1)
{$\tilde{\mathbf{H}}_t = \mathbf{H}_t + \mathbf{P}$};

\draw (2.5,-3) -- (2.5+\W,-3);
\draw (2.5,-2) -- (2.5+\W,-2);
\draw (3.5,-4) -- (3.5,-1);
\draw (4.5,-4) -- (4.5,-1);

\node at (3,-1.5) {$\tilde{h}_{t-L+1}^{(1)}$};
\node at (4,-1.5) {$\cdots$};
\node at (5,-1.5) {$\tilde{h}_{t}^{(1)}$};

\node at (3,-2.5) {$\vdots$};
\node at (4,-2.5) {$\ddots$};
\node at (5,-2.5) {$\vdots$};

\node at (3,-3.5) {$\tilde{h}_{t-L+1}^{(d)}$};
\node at (4,-3.5) {$\cdots$};
\node at (5,-3.5) {$\tilde{h}_{t}^{(d)}$};

\end{tikzpicture}
\caption{
Positional encoding. The embedding matrix $\mathbf{H}_t$ is augmented by the deterministic positional encoding matrix $\mathbf{P}$ through element-wise addition, producing $\tilde{\mathbf{H}}_t$ while preserving dimensionality.
}
\label{fig:positional_encoding}
\end{figure}
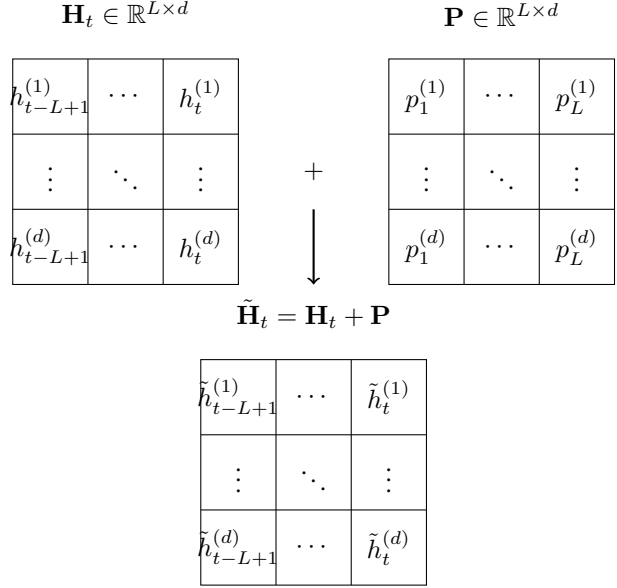

Let the projected feature matrix be denoted by $\mathbf{H}_t \in \mathbb{R}^{L \times d}$. A positional encoding matrix $\mathbf{P} \in \mathbb{R}^{L \times d}$ is constructed with entries defined by
\[
P_{\tau,2k} = \sin\!\left(\frac{\tau}{10000^{2k/d}}\right), 
\qquad
P_{\tau,2k+1} = \cos\!\left(\frac{\tau}{10000^{2k/d}}\right),
\]
for $\tau = 0,\ldots,L-1$ and $k = 0,\ldots,\lfloor d/2 \rfloor - 1$. As illustrated in Figure~\ref{fig:positional_encoding}, the position-aware embedding is obtained via elementwise addition,
\[
\widetilde{\mathbf{H}}_t = \mathbf{H}_t + \mathbf{P}.
\]
This operation preserves dimensionality while injecting structured temporal information into each feature vector. The multi-scale sinusoidal construction encodes both absolute and relative temporal position without introducing additional trainable parameters. In particular, different frequency components allow the model to infer relative distances between time steps while maintaining absolute ordering within the window. The enriched representation $\widetilde{\mathbf{H}}_t$ is subsequently processed by the transformer encoder to capture long-range temporal interactions across the full window.

\noindent Subsequently, a transformer encoder with multi-head self-attention is applied to the position-enhanced embedding sequence $\widetilde{\mathbf{H}}_t$ to model global temporal dependencies. The self-attention mechanism allows each time index within the window to attend to all others, thereby capturing long-range interactions and regime-level structure that may not be locally observable. The transformer output remains a sequence in $\mathbb{R}^{L \times d}$, preserving temporal resolution.

\noindent In parallel, a bidirectional LSTM processes the same position-enhanced embedding sequence to capture sequential and directional dynamics. Whereas self-attention emphasizes global relational structure, the recurrent branch explicitly models ordered temporal propagation in both forward and backward directions, providing sensitivity to persistence, momentum, and transient sequential effects. Let the pooled outputs of the transformer and recurrent branches be denoted by 
\[
\mathbf{h}_t^{(\mathrm{attn})} \in \mathbb{R}^{d_a}, 
\qquad 
\mathbf{h}_t^{(\mathrm{rnn})} \in \mathbb{R}^{d_r},
\]
where $d_a$ and $d_r$ denote the corresponding summary dimensions after global pooling. These representations are concatenated to form
\[
\mathbf{h}_t = 
\big[
\mathbf{h}_t^{(\mathrm{attn})} ;
\mathbf{h}_t^{(\mathrm{rnn})}
\big]
\in \mathbb{R}^{d_a + d_r},
\]
and mapped through a fully connected layer to obtain a latent vector
\[
\mathbf{z}_t \in \mathbb{R}^{q},
\]
which serves as a compact embedding of both local convolutional structure and global temporal dependencies within the input window.

To enhance robustness under regime changes and potential model misspecification, a second-stage refinement mechanism is introduced. Let 
\[
\widehat{\mathbf{F}}_{t,1} \in \mathbb{R}^{H \times p}
\]
denote the initial forecast generated from the latent representation $\mathbf{z}_t$. The corresponding forecast residual is defined as
\[
\mathbf{R}_t
=
\mathbf{F}_t - \widehat{\mathbf{F}}_{t,1}.
\]
The residual matrix is vectorized and concatenated with $\mathbf{z}_t$, and the resulting representation is passed through a refinement network to produce a corrected forecast
\[
\widehat{\mathbf{F}}_{t,2} \in \mathbb{R}^{H \times p}.
\]
Although the residual matrix $\mathbf{R}_t \in \mathbb{R}^{H \times p}$ is vectorized before entering the refinement network, the dimensionality scales linearly in the cross-sectional dimension $p$. Because the forecast horizon $H$ is typically modest in multi-step prediction settings, the refinement stage introduces only a single additional projection and does not materially increase computational complexity relative to the backbone architecture.

This two-pass architecture enables the refinement stage to model structured residual components not captured by the initial forecast, thereby improving robustness to structural shifts and heterogeneous temporal dynamics. Let $w_1, w_2, w_r \ge 0$ denote weights governing the contributions of the first-pass forecast loss, refined forecast loss, and reconstruction loss, respectively, and let $\lambda \ge 0$ denote a latent regularization parameter. The model parameters are obtained by minimizing the composite objective
\[
\mathcal{L}
=
w_1 \,
\|\mathbf{F}_t - \widehat{\mathbf{F}}_{t,1}\|_F^2
+
w_2 \,
\|\mathbf{F}_t - \widehat{\mathbf{F}}_{t,2}\|_F^2
+
w_r \,
\|\mathbf{X}_t - \widehat{\mathbf{X}}_t\|_F^2
+
\lambda \,
\|\mathbf{z}_t\|_2^2,
\]
where $\|\cdot\|_F$ denotes the Frobenius norm. The final term regularizes the latent representation to control variance and mitigate over-dispersion in the embedding space. In practice, the weights are selected to place greater emphasis on the refined forecasting objective while preserving reconstruction fidelity.

\subsection{Reconstruction-Based Purification}

Let $\{\mathbf{X}_t\}_{t=L}^{T-H}$ denote the collection of rolling windows constructed from the full time series $\{\mathbf{x}_t\}_{t=1}^T$. In unsupervised anomaly detection, these windows may themselves be contaminated by anomalous or structurally shifted behavior. If such windows are used directly to estimate the generative backbone, the learned representation may partially internalize abnormal dynamics, thereby weakening subsequent anomaly discrimination.

To obtain a more robust estimate of baseline system behavior, we introduce a preliminary purification stage. First, the backbone architecture is fitted using reconstruction loss only,
\[
\mathcal{L}_{\mathrm{recon}}
=
\|\mathbf{X}_t - \widehat{\mathbf{X}}_t\|_F^2,
\]
for all windows in an initial index set $\mathcal{I}_0 = \{L,\dots,T-H\}$. That is, during purification the forecasting and refinement components are disabled and the model parameters are estimated solely by minimizing $\mathcal{L}_{\mathrm{recon}}$. This produces reconstruction errors
\[
e_t
=
\|\mathbf{X}_t - \widehat{\mathbf{X}}_t\|_F^2.
\] Let $q_{1-\alpha}$ denote the empirical $(1-\alpha)$-quantile of $\{e_t : t \in \mathcal{I}_0\}$ for a small trimming parameter $\alpha \in (0,1)$. Candidate contaminated windows are defined as
\[
\mathcal{A}^{(1)}
=
\{ t \in \mathcal{I}_0 : e_t \ge q_{1-\alpha} \}.
\]
To prevent excessive removal in heavy-tailed or highly volatile environments, the fraction of discarded windows is capped at a pre-specified level $\gamma_{\max}$. The purified index set is then defined as
\[
\mathcal{I}_1
=
\mathcal{I}_0 \setminus \mathcal{A}^{(1)}.
\] The procedure may be iterated: at iteration $k$, the model is re-estimated on $\mathcal{I}_{k-1}$, reconstruction errors are recomputed, and a new contaminated set $\mathcal{A}^{(k)}$ is identified using the same quantile-based rule. The iterations terminate when the index set stabilizes or after a fixed number of rounds. The final backbone model is then estimated using the purified window set $\mathcal{I}_K$.

\subsection{Ensemble Anomaly Scoring}

Anomalous behavior in high-dimensional time series may manifest through multiple mechanisms, including forecast breakdown, reconstruction instability, latent-space sparsity, and shifts in dependence structure. Reliance on a single diagnostic is therefore often insufficient. To address this heterogeneity, ReGEN-TAD constructs a composite anomaly score by aggregating multiple complementary window-level statistics. For each time index $t$, let $\mathcal{S}_t = \{ s_t^{(m)} \}_{m=1}^M$ denote a collection of diagnostic scores derived from the generative backbone. In the current implementation, $M=6$, corresponding to the  component statistics:
\[
\begin{array}{l}
s_t^{(1)} : \text{Refined forecasting residual magnitude} \\
s_t^{(2)} : \text{Reconstruction residual magnitude} \\
s_t^{(3)} : k\text{-nearest-neighbor latent density score} \\
s_t^{(4)} : \text{Latent residual dynamics deviation} \\
s_t^{(5)} : \text{Mahalanobis distance from baseline latent distribution} \\
s_t^{(6)} : \text{Dispersion of forecast residuals}
\end{array}
\]
The Mahalanobis component $s_t^{(5)}$ requires estimation of the
$q$-dimensional latent covariance. To avoid instability
from naive inversion, we employ a shrinkage precision estimator
(Ledoit--Wolf), with a small ridge
regularization term to ensure well-conditioned inversion. Formally, each component score may be written as
\[
s_t^{(m)} = \phi_m(\mathbf{X}_t, \mathbf{F}_t, \mathbf{z}_t),
\]
where $\phi_m$ denotes a measurable functional capturing a distinct structural dimension of abnormality. To ensure numerical stability when the interquartile range is close to zero, we introduce a small positive constant $\epsilon > 0$ and compute
\[
\widetilde{s}_t^{(m)}
=
\frac{|s_t^{(m)} - \mathrm{med}(s^{(m)})|}
{\mathrm{IQR}(s^{(m)}) + \epsilon},
\]
where $\epsilon = 10^{-6}$ in all experiments. This robust standardization places the heterogeneous diagnostics on a comparable scale while limiting sensitivity to extreme observations. The overall anomaly score is then defined as a weighted aggregation,
\[
S_t
=
\frac{1}{M}
\sum_{m=1}^M
w_m \, \widetilde{s}_t^{(m)},
\]
where $w_m \ge 0$ are user-specified or data-driven weights satisfying $\sum_{m=1}^M w_m = M$ for scale normalization. Importantly, the ensemble structure is modular: the collection $\{\phi_m\}_{m=1}^M$ is not fixed by design. Depending on the application, additional diagnostics may be incorporated or existing components removed without altering the backbone architecture. In addition, because scoring occurs in a lower intrinsic latent manifold, instability in high-dimensional $k$-nearest-neighbor and other scorers is attenuated.

\subsection{Score Calibration and Decision Rules}

Given the ensemble anomaly score $S_t$, the final step consists of mapping scores to a binary anomaly indicator. Let
\[
\widehat{A}_t \in \{0,1\}
\]
denote the predicted anomaly label at time index $t$, where $\widehat{A}_t = 1$ indicates anomalous behavior.

\paragraph{Threshold-based detection.}
Given the ensemble anomaly score $S_t$, detection is performed using a fixed global quantile threshold. Specifically, let $\alpha \in (0,1)$ denote the contamination parameter. The threshold is defined as
\[
\tau
=
\mathrm{Quantile}_{1-\alpha}(S_1,\dots,S_N),
\]
and windows are flagged according to
\[
\widehat{A}_t
=
\mathbf{1}\{ S_t > \tau_\alpha \}.
\]
This approach fixes the global anomaly rate at $\pi$ and avoids time-varying normalization effects that may attenuate sensitivity during concentrated regime shifts.

\paragraph{Rank-based detection.}
To mitigate this effect, we additionally consider a rank-based decision rule. Let $N$ denote the number of evaluated windows and let $\alpha \in (0,1)$ be a user-specified anomaly proportion. The rank-based detector flags the top $\lceil \alpha N \rceil$ windows with largest scores:
\[
\widehat{A}_t
=
\mathbf{1}\{ S_t \text{ is among the top } \alpha \text{ fraction of scores} \}.
\]
This approach is particularly effective in settings with sparse but clustered anomalies. To promote temporal coherence short isolated detections may be removed via a minimum run-length filter, and detected segments may be dilated by a small neighborhood radius to capture contiguous abnormal regimes.

\subsection{Hyperparameter Configuration and Sensitivity Analysis}

The default hyperparameter configuration adopted throughout this study is not ad hoc, but rather the result of a systematic sensitivity analysis over the principal architectural and scoring components of the proposed framework. The objective of this analysis was to identify a parameter regime that provides stable and robust performance across heterogeneous datasets, varying dimensionality, and differing anomaly densities. Sensitivity experiments over embedding dimension, attention width, LSTM units, purification quantile, latent neighbor count, ensemble weights, and smoothing span revealed gradual rather than abrupt performance degradation, indicating that the proposed framework does not rely on narrowly tuned hyperparameters. While additional gains may be achievable through problem-specific tuning, the selected configuration is intended to provide reliable performance across diverse financial time-series settings without dataset-specific optimization.

\paragraph{Backbone architecture.}
The convolutional front-end consists of two temporal convolutional layers with 64 filters and kernel width three, followed by projection into an embedding space of dimension $d=128$. The transformer encoder employs six attention heads with feed-forward width 128 and dropout rate 0.1. In parallel, a bidirectional LSTM with 32 hidden units per direction captures sequential dynamics. The pooled outputs of the attention and recurrent branches are concatenated and mapped to a latent representation of dimension $q=128$ prior to the forecasting and reconstruction heads.

\paragraph{Training configuration.}
Model parameters are optimized using the Adam algorithm with learning rate $10^{-3}$. The composite objective emphasizes refined forecasting accuracy while preserving reconstruction fidelity, with weights $(w_1, w_2, w_r) = (0.2, 0.8, 0.5)$ assigned to the initial forecast, refined forecast, and reconstruction terms, respectively. Latent $\ell_2$ regularization is set to $\lambda=0$ in the baseline configuration, although mild regularization was examined during sensitivity analysis.

\begin{algorithm}[!t]
\caption{ReGEN-TAD Algorithm}
\label{algo:regenad}
\begin{algorithmic}[1]
\Require $\{\mathbf{x}_t\}_{t=1}^T$, split $T_0<T_1$, $L,H$, weights $\mathbf{w}$, tail $\alpha$, purif. quant. $q$
\Ensure Scores $\{s_t\}$, labels $\{\hat a_t\}$

\Statex \textbf{Split:}
\State $\mathcal{I}_{tr}=\{L,\dots,T_0-H\}$, $\mathcal{I}_{cal}=\{T_0-H+1,\dots,T_1-H\}$, $\mathcal{I}_{te}=\{T_1-H+1,\dots,T-H\}$

\Statex \textbf{STAGE I: Reconstruction-Based Outlier Purification}
\State $\mathcal{I} \gets \mathcal{I}_{tr}$
\Repeat
    \State Fit $\mathcal{G}_0$ on $\mathcal{I}$ minimizing $e_t = \|\mathbf{X}_t - \hat{\mathbf{X}}_t\|_F^2$
    \State $\mathcal{I} \gets \{t \in \mathcal{I} : e_t < Q_{q}(\{e_s : s \in \mathcal{I}\})\}$
\Until{converged}

\Statex \textbf{STAGE II: Generative Backbone Training}
\State Fit backbone $\mathcal{G}$ on purified set $\mathcal{I}$ minimizing:
\State $\mathcal{L} = w_1\|F_t - \hat{F}_{t,1}\|_F^2 + w_2\|F_t - \hat{F}_{t,2}\|_F^2 + w_r\|X_t - \hat{X}_t\|_F^2 + \lambda\|z_t\|_2^2$

\Statex \textbf{Calibration (on $\mathcal{I}_{cal}$ only)}
\State Compute component scores $s_t^{(m)}$
\State Fit $(\text{med}_m, \text{IQR}_m)$ for each $m$
\State Aggregate $S_t$ on $\mathcal{I}_{cal}$ and set cutoff \\
$\tau_\alpha = Q_{1-\alpha}\!\left(\{ S_t : t \in \mathcal{I}_{cal} \}\right)$

\Statex \textbf{STAGE III: Ensemble Scoring and Decision Function}
\For{$t \in \mathcal{I}_{te}$}
    \State Standardize $\tilde s_t^{(m)} = \frac{|s_t^{(m)} - \text{med}_m|}{\text{IQR}_m + \epsilon}$, where $\epsilon = 10^{-6}$
    \State $S_t \gets \mathrm{EWMA}\!\left( \sum_m w_m \widetilde{s}_t^{(m)} \right)$
    \State $\widehat{A}_t \gets \mathbf{1}\{ S_t \ge \tau_\alpha \}$
\EndFor
\State \Return $\{\hat a_t\}, \{s_t\}$
\end{algorithmic}
\end{algorithm}

\paragraph{Purification stage.}
The purification mechanism trims candidate anomalous windows using a high reconstruction-error quantile threshold ($q=0.97$) and a maximum removal fraction of 30\%. Moderate deviations from these values produce limited changes in detection performance, suggesting that the procedure is not critically dependent on fine calibration.

\paragraph{Ensemble scoring and calibration.}
The anomaly score aggregates multiple diagnostic components, including refined forecast error, reconstruction error, $k$-nearest-neighbor latent distance (with $k=20$), latent dynamic deviation (lag 5), Mahalanobis regime distance, and residual volatility. Each component is robustly normalized using median and interquartile range statistics and combined via weighted averaging. The anomaly fraction parameter is set to $\alpha=0.05$, and temporal smoothing is performed using an EWMA filter with span five. The decision mechanism defaults to a rank-based rule that flags approximately the top 5\% of windows, although a threshold-based alternative with adaptive rolling statistics is also supported.

\begin{figure*}[t]
\centering
\includegraphics[width=\textwidth]{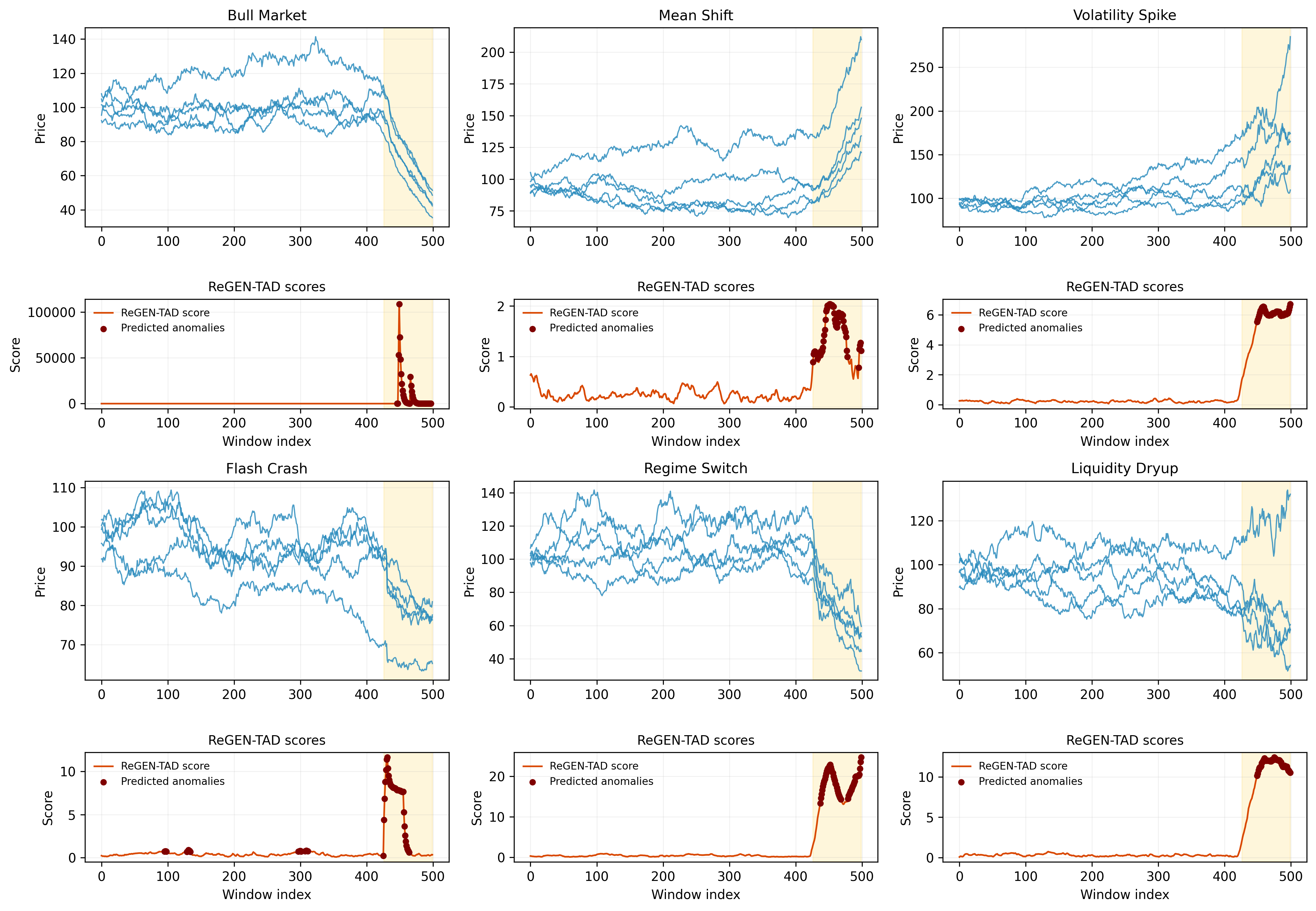}
\caption{
Illustration of ReGEN-TAD across six synthetic financial regimes:
Bull Market, Mean Shift, Volatility Spike, Flash Crash, Regime Switch, and Liquidity Dryup.
Top panels show multivariate price trajectories (sample of first 5 series) with anomaly region shaded.
Bottom panels show corresponding ReGEN-TAD anomaly scores and predicted anomalies.
}
\label{fig:regentad_scenarios}
\end{figure*}

\paragraph{Illustrative example.}
Figure~\ref{fig:regentad_scenarios} demonstrates that, under a fixed hyperparameter configuration, ReGEN-TAD produces stable scores during in-control periods and sharp, well-localized detections across diverse anomaly mechanisms. These behaviors are obtained without scenario-specific tuning, illustrating robustness across heterogeneous anomaly mechanisms.

\begin{figure*}[t]
\centering
\includegraphics[width=\textwidth]{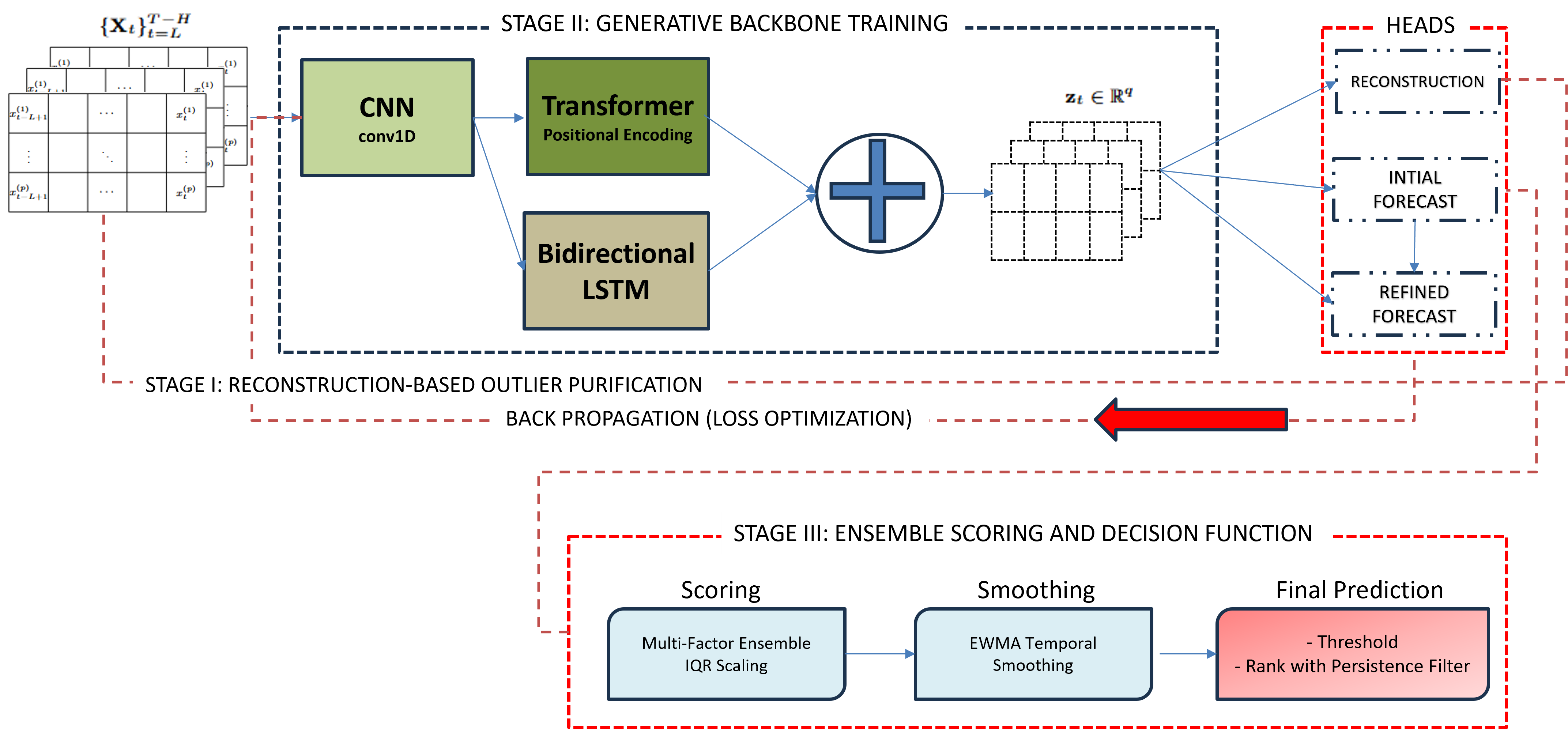}
\caption{
Overall architecture of ReGEN-TAD. The generative backbone transforms input windows through convolutional, positional, attention-based, and recurrent encoding to produce a latent representation. Forecasting and reconstruction heads generate diagnostic signals, which are aggregated in the decision function using robust scaling and smoothing before applying rank- or threshold-based anomaly selection.
}
\label{fig:regenad_architecture}
\end{figure*}

\subsection{Overall Architecture and Algorithmic Structure}

Algorithm~\ref{algo:regenad} formalizes the complete three-stage procedure underlying the proposed ReGEN-TAD framework. The algorithm makes explicit the separation between contamination control, representation learning, and anomaly calibration, which are implemented sequentially but remain modular. This structured decomposition clarifies the role of each component and facilitates transparent analysis of how instability is isolated, represented, and ultimately classified within the unified framework.

Stage~I performs reconstruction-based purification to mitigate contamination in the reference set. An auxiliary generative model is trained using reconstruction loss only, and windows exhibiting extreme reconstruction error are iteratively removed subject to a trimming constraint. This yields a purified index set intended to approximate baseline behavior, thereby stabilizing subsequent representation learning and reducing the risk that anomalous structure is internalized during model estimation.

Stage~II trains the generative backbone on the purified data. Each sliding window $\mathbf{X}_t$ is processed through a convolutional feature extractor, augmented with positional encoding, and passed through parallel transformer and bidirectional recurrent branches. The pooled outputs are concatenated and mapped to a latent embedding $\mathbf{z}_t$, which captures both local and global temporal structure. Forecasting and reconstruction heads operate on $\mathbf{z}_t$, and a second-stage refinement network produces an adjusted forecast to model systematic residual structure beyond the initial prediction.

Stage~III converts model outputs into anomaly decisions. Multiple diagnostic components are constructed from forecast residuals, reconstruction deviation, latent-space distances, dynamic instability measures, and volatility features. These components are robustly standardized using median and interquartile range statistics, aggregated into a unified anomaly score, and temporally smoothed via EWMA filtering. Final anomaly labels are obtained either by selecting the top-$\alpha$ fraction of windows (rank-based mode) or by comparing the score to a calibrated threshold.

Figure~\ref{fig:regenad_architecture} provides a schematic representation of the three-stage architecture, illustrating the flow from rolling-window inputs through the generative backbone to the ensemble scoring and decision layer. This structured design improves interpretability, clarifies the interaction between stages, and enhances robustness across heterogeneous anomaly mechanisms.

\subsection{Factor Attribution and Interpretability}

A central objective of ReGEN-TAD is not only accurate anomaly detection, but economically meaningful interpretation. To this end, the framework provides factor-level attribution by decomposing each detected anomaly into contributions that reflect both statistical abnormality and model-driven relevance. Let $\mathbf{X}_t \in \mathbb{R}^{L \times p}$ denote an anomalous window, where $L$ is the window length and $p$ the number of factors. The goal is to identify which factors simultaneously deviate from baseline behavior and materially influence the model's anomaly assessment, thereby linking representation learning to economically interpretable structure.

\vspace{0.5em}
\noindent\textbf{Baseline deviation.}
For each factor $j$, we quantify standardized deviation from its baseline regime:
\begin{equation}
\Delta_j
=
\frac{\lvert \bar{x}_{t,j} - \mu_j \rvert}{\sigma_j},
\end{equation}
where $\bar{x}_{t,j} = L^{-1}\sum_{\ell=1}^L x_{\ell,j}$ is the window average and $(\mu_j,\sigma_j)$ are baseline estimates from calibration. This term measures economic abnormality in volatility-scaled units and captures departures from typical cross-sectional behavior. In financial terms, $\Delta_j$ isolates factors experiencing unusually large shifts relative to their historical dispersion.

\vspace{0.5em}
\noindent\textbf{Latent sensitivity.}
To assess model relevance, define the latent anomaly score
\[
\mathcal{S}(\mathbf{X}_t)
=
\| z(\mathbf{X}_t) - \mu_z \|^2,
\]
where $z(\mathbf{X}_t)$ is the latent embedding and $\mu_z$ its baseline center. Factor-level sensitivity is
\begin{equation}
\Gamma_j
=
\frac{1}{L}
\sum_{\ell=1}^L
\left|
\frac{\partial \mathcal{S}(\mathbf{X}_t)}{\partial x_{\ell,j}}
\right|.
\end{equation}
Gradients are computed over the entire window so that $\Gamma_j$ reflects how perturbations in factor $j$ propagate through the learned temporal representation, capturing systemic influence rather than isolated pointwise effects. This term therefore quantifies how strongly movements in factor $j$ alter the internal anomaly representation learned by the generative backbone.

\vspace{0.5em}
\noindent\textbf{Factor attribution.}
The contribution of factor $j$ is defined as
\begin{equation}
C_j
=
\Delta_j \Gamma_j,
\end{equation}
optionally normalized so that $\sum_{j=1}^p C_j = 1$. The multiplicative structure ensures that high attribution is assigned only to factors that are both economically abnormal and structurally influential in the learned representation. Factors are ranked by $C_j$, and a cumulative-mass rule identifies the smallest subset explaining a specified proportion of total contribution, yielding sparse and economically interpretable explanations. This decomposition bridges deep generative modeling with factor-based economic intuition, aligning anomaly detection with the interpretability standards of financial econometrics.

\begin{figure*}[!tb]
\centering
\includegraphics[width=\textwidth]{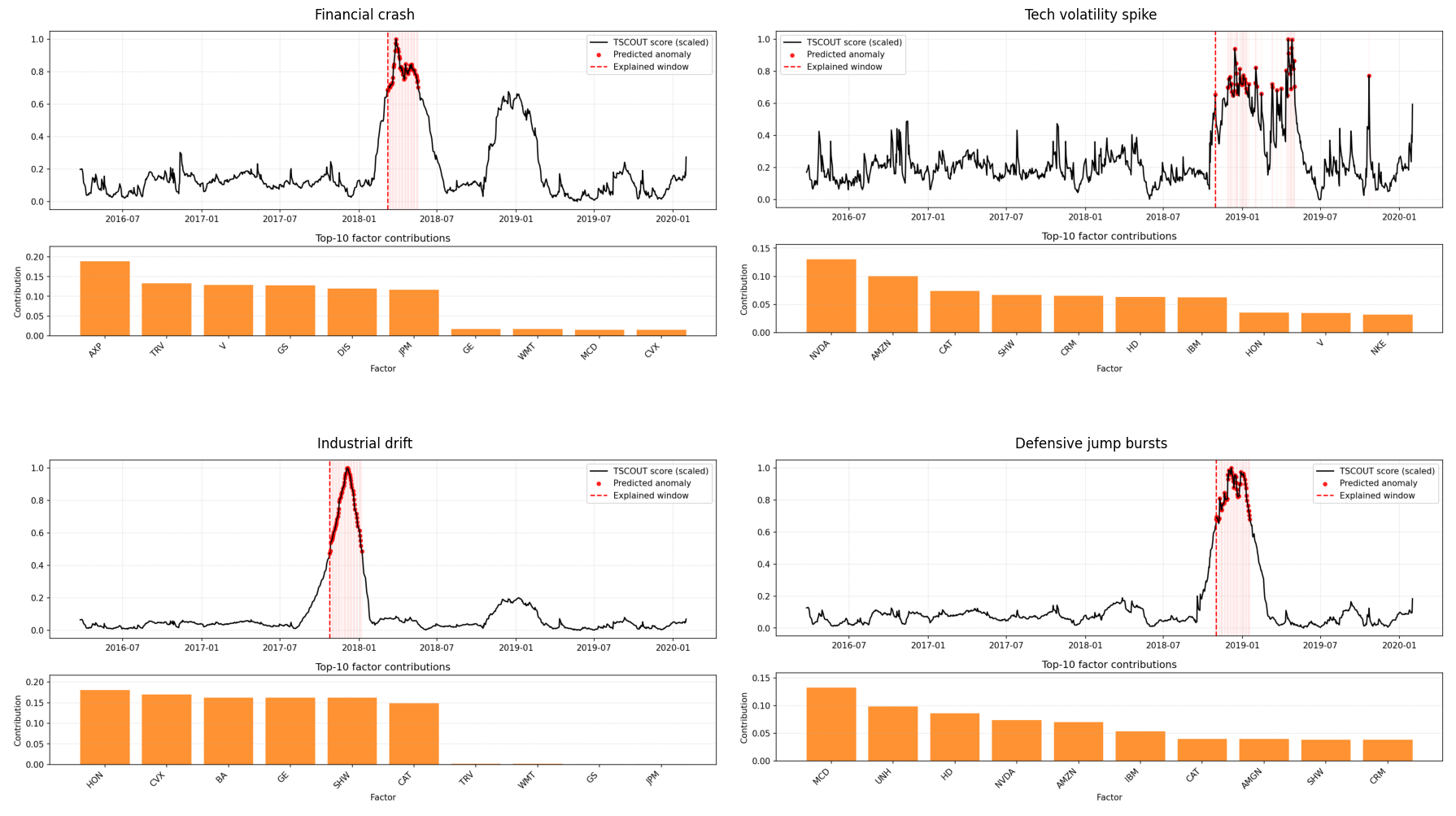}
\caption{%
Factor-level attribution for sector-specific synthetic anomalies detected by ReGEN-TAD. 
Each panel corresponds to a distinct anomaly type injected into a single economic sector (financials, technology, industrials, and defensive assets). 
For each anomaly, the top subpanel displays the normalized ReGEN-TAD anomaly score over time, with detected anomalous windows highlighted, while the bottom subpanel reports the top contributing factors for the explained window based on standardized deviation from the baseline and reconstruction sensitivity. 
The results illustrate ReGEN-TAD's ability to localize anomalies temporally and attribute them to economically relevant factors.}
\label{fig:attribution}
\end{figure*}

Figure~\ref{fig:attribution} illustrates the sector-localization capability of ReGEN-TAD under controlled anomaly injections. In these experiments, anomalies are deliberately introduced into specific sectors: financials, technology, industrials, and defensive stocks, so that the ground-truth source of abnormality is known ex ante. The attribution analysis therefore serves as a validation exercise: if the interpretability mechanism is effective, the highest contributions should concentrate on the factors belonging to the perturbed sector.

In the financial crash scenario, the anomaly is injected directly into financial stocks, and the dominant contributions are correspondingly assigned to financial institutions (e.g., AXP, GS, JPM), reflecting concentrated stress within that sector rather than diffuse cross-sectional noise. Technology-driven volatility spikes are introduced exclusively in large-cap technology names, and attribution mass concentrates on NVDA, AMZN, and related firms. The industrial drift experiment imposes a systematic downward trend on industrial and cyclical stocks, and the resulting attribution profile loads primarily on CAT, BA, HON, and related industrial firms. Similarly, defensive jump bursts are injected into traditionally defensive assets, and the highest contributions are allocated to those same names.

Across scenarios, attribution remains sparse, sector-coherent, and closely aligned with the known injection structure. This controlled design demonstrates that the framework does not merely amplify high-variance components or mechanically select large-cap factors; rather, it correctly identifies the economic origin of the anomaly. The experiment therefore provides direct evidence that the proposed gradient-based attribution mechanism recovers meaningful drivers of instability, linking deep generative representations to interpretable sector-level economic structure. Importantly, because the injected shock location is known by construction, the alignment between attribution mass and the perturbed sector serves as an explicit validation of interpretability.

\section{Simulation Studies}
\label{sec:experiments}

We organize the empirical evaluation into four complementary environments designed to stress distinct aspects of the proposed framework: general structural anomalies, market-regime and cross-sectional shock simulations, sensitivity to prediction horizon, and specificity under clean regimes. This design separates purely statistical detection behavior from economically structured stress scenarios and allows us to evaluate robustness, interpretability, and calibration stability in a unified setting.

Across all experiments, ReGEN-TAD is compared against a diverse set of baselines spanning reconstruction-based, forecasting-based, density-based, transformer-based, and classical econometric paradigms. Deep generative and forecasting competitors include DAGMM \citep{zong2018deep}, which combines autoencoding with Gaussian mixture density estimation; DeepAnt \citep{munir2018deepant}, a convolutional forecasting detector; TranAD \citep{tuli2022tranad}, a transformer-based adversarial reconstruction model; and TGAN-AD, a generative adversarial time-series detector. We additionally include the transformer-based positional encoding framework (AlioghliOkay) of \citet{alioghli2025enhancing}, which enhances multivariate time-series anomaly detection through structured temporal encoding mechanisms, and TimeGPT-1 \citep{garza2023timegpt}, a large-scale pretrained model for time-series forecasting accessed via commercial API and adapted for anomaly scoring. Because TimeGPT is accessed via API and relies on external pretrained weights, we report results obtained under fixed query specifications without fine-tuning. Isolation Forest \citep{liu2008isolation} is also incorporated as a strong tree-based ensemble widely used in high-dimensional anomaly detection.

We also incorporate strong linear and structural baselines popular in financial econometrics. These include multivariate residual-based anomaly detection using ordinary least squares (OLS) and reduced-rank regression (RRR) residual scoring to capture low-rank cross-sectional dependence structures. GARCH-based volatility monitoring is included to represent conditional heteroskedastic modeling standard in empirical asset pricing and risk management. By benchmarking against both modern deep-learning architectures and established econometric instability diagnostics, the evaluation ensures that performance improvements reflect genuine structural advantages rather than reliance on a specific modeling paradigm.

All neural-network-based models are implemented using architectures consistent with their original publications and are trained under a similar optimization protocol. Training is conducted with identical data splits, rolling windows, optimizer, learning rate schedule, batch size, and epochs. No model receives oracle access to anomaly labels, and no dataset-specific hyperparameter tuning is performed beyond predefined calibration. Classical econometric baselines are estimated using the same rolling windows and without access to future information.

We adopt a leakage-free rolling calibration protocol: training on $[1,T_0]$, validation on $[T_0+1,T_1]$, and testing on $[T_1+1,T]$. Purification and median/IQR normalization on ReGEN-TAD are computed using training data only; thresholds are fixed on validation data and remain unchanged during testing. Normalization statistics for ReGEN-TAD (Algorithm~1, Step~10) are computed exclusively from past information, preventing forward-looking contamination.

Performance is evaluated using Precision, Recall, F1-score, False Positive Rate (FPR), AUROC, and runtime. Runtime is measured as wall-clock training and inference time (seconds) under identical conditions on an Intel Core i9-7980XE CPU and NVIDIA RTX 3090 GPU, providing a practical comparison rather than formal complexity analysis.

Taken together, this evaluation framework provides a rigorous comparison across heterogeneous modeling paradigms under economically realistic and statistically controlled environments while maintaining strict out-of-sample integrity. While numerous additional anomaly detection and structural-break procedures exist, it is infeasible to benchmark against all variants; the selected comparators were chosen to span the principal methodological paradigms, including deep generative, forecasting-based, transformer, density-based, and classical econometric residual monitors, thereby providing representative coverage of the dominant approaches in both machine learning and financial econometrics.

\subsection{General Structural Anomalies}

We begin by evaluating performance under controlled structural anomaly mechanisms embedded in high-dimensional stochastic processes. This environment isolates fundamental detection capabilities under clearly defined contamination regimes while preserving realistic cross-sectional dependence and temporal dynamics characteristic of financial return panels. By abstracting from market microstructure and firm-specific heterogeneity, this setting provides a clean laboratory for understanding how generative modeling, purification, and ensemble aggregation interact under well-characterized statistical shifts.

Multivariate series $\{\mathbf{x}_t\}_{t=1}^T$, with $\mathbf{x}_t \in \mathbb{R}^p$, are generated from stationary autoregressive processes with structured cross-sectional covariance and optional seasonal components. To better reflect empirical return behavior, innovations are drawn from both Gaussian and heavy-tailed distributions. We examine dimension $p \in \{100\}$. All structural simulations are generated with total length T=500; results are computed over rolling windows determined by the specified window length and forecast horizon. All series are expressed in returns to mitigate scale distortions and spurious nonstationarity. Autoregressive dependence ensures that anomalies propagate temporally rather than appearing as independent outliers, reflecting persistence patterns typical of economic systems.

Structural anomalies are injected after baseline generation using predefined mechanisms. We consider persistent mean shifts, variance shifts, deterministic trend deviations, transient spikes, collective cross-sectional disruptions affecting subsets of variables, and contextual anomalies whose abnormality depends on prevailing state conditions. The contamination proportion $\gamma$ governs anomaly duration, with low-contamination regimes $\gamma \in \{0.01,0.03,0.05\}$ and high-contamination regimes $\gamma \in \{0.10,0.12,0.15\}$. To assess robustness to contaminated training segments, anomalies are introduced either early or late in the sample. This design directly evaluates the stability of the purification and rolling calibration stages when portions of the estimation window contain mild structural shifts. Across all anomaly types, contamination regimes, and injection timings, performance is averaged over 3,900 independent Monte Carlo replications, ensuring statistical stability and preventing conclusions from being driven by isolated realizations.

\vspace{0.5em}
\noindent\textbf{Trend shifts.}
Trend anomalies introduce deterministic drift into the data-generating process. A representative formulation is
\begin{equation}
\mathbf{Y}_{t+k}
=
\boldsymbol{\mu}
+
k\,\boldsymbol{\beta}
+
\boldsymbol{\varepsilon}_{t+k},
\qquad k \geq 0,
\end{equation}
where $\boldsymbol{\beta} \neq \mathbf{0}$ induces gradual deviation from the baseline trajectory. Unlike abrupt shocks, trend shifts accumulate linearly over time and manifest as persistent forecast bias. In forecasting-based detectors, the cumulative discrepancy
\[
\max_{1 \leq k \leq h}
\mathcal{D}(\hat{\mathbf{Y}}_{t+k-1}, \mathbf{Y}_{t+k-1})
\]
tends to increase with the horizon $h$, since prediction errors compound as the drift evolves. Consequently, longer horizons can amplify the anomaly signal. In simulations, slopes are set to $\beta_j = 0.05$ for affected dimensions and $\beta_j = 0$ otherwise.

\vspace{0.5em}
\noindent\textbf{Spike anomalies.}
Spike anomalies correspond to isolated, high-magnitude deviations:
\begin{equation}
\mathbf{Y}_{t+k}
=
\boldsymbol{\mu}
+
\boldsymbol{\delta}\,\mathbb{I}(k = k^\star)
+
\boldsymbol{\varepsilon}_{t+k},
\end{equation}
where $k^\star$ denotes a single affected time index and $\boldsymbol{\delta} \neq \mathbf{0}$ specifies the spike magnitude. These events produce sharp but short-lived forecast residuals. Because the detection rule often depends on the maximum discrepancy within a horizon window, extending $h$ beyond the spike location does not introduce additional information once the spike is captured. Thus, detection probability saturates after inclusion of the extreme deviation. In simulations, we set $\delta_j = 3\sigma$ for affected dimensions and $\delta_j = 0$ otherwise, yielding pronounced but transient shocks.

\vspace{0.5em}
\noindent\textbf{Mean shifts.}
Persistent mean shifts reflect regime changes in expected returns:
\begin{equation}
\mathbf{Y}_{t+k}
=
\boldsymbol{\mu}
+
\boldsymbol{\Delta}
+
\boldsymbol{\varepsilon}_{t+k},
\qquad
\boldsymbol{\Delta} \neq \mathbf{0}.
\end{equation}
Here, the deviation is constant rather than evolving. Forecasting models may partially adapt over time, reducing prediction error as the new regime becomes incorporated into recent observations. In simulations, we set $\Delta_j = 1.5\sigma$ for affected dimensions and $\Delta_j = 0$ otherwise.

\vspace{0.5em}
\noindent\textbf{Variance shifts.}
Variance anomalies alter the second-order structure of the process without necessarily shifting the mean:
\begin{equation}
\mathbf{Y}_{t+k}
\sim
\mathcal{N}(\boldsymbol{\mu},\, \boldsymbol{\Sigma}_1),
\qquad
\boldsymbol{\Sigma}_1 \neq \boldsymbol{\Sigma}_0.
\end{equation}
In simulations, we inflate variances for affected dimensions by a factor of two, setting $\Sigma_{1,jj} = 2\sigma^2$ while leaving other entries unchanged.

\vspace{0.5em}
\noindent\textbf{Collective anomalies.}
Collective anomalies affect coherent subsets of variables simultaneously, reflecting cross-sectional stress propagation. Let $\mathcal{J} \subset \{1,\dots,p\}$ denote an affected index set. Then
\[
x_{t,j}
=
\begin{cases}
\mu_j + \Delta_j + \varepsilon_{t,j}, & j \in \mathcal{J}, \\
\mu_j + \varepsilon_{t,j}, & j \notin \mathcal{J}.
\end{cases}
\]
Such deviations preserve internal structure within the affected group but alter global covariance patterns. These anomalies are particularly challenging for methods that treat dimensions independently. In simulations, we set $|\mathcal{J}| = 0.25p$ and $\Delta_j = 1.5\sigma$ for $j \in \mathcal{J}$.

\vspace{0.5em}
\noindent\textbf{Contextual anomalies.}
Contextual anomalies depend on prevailing states rather than absolute magnitudes. A stylized representation is
\[
x_{t,j}
=
\mu_j
+
\Delta_j \,\mathbb{I}(x_{t-1,j} > 0)
+
\varepsilon_{t,j},
\]
so that abnormality is conditional on the sign or regime of the process. Such anomalies cannot be identified solely through marginal distributional changes; instead, they require modeling of temporal dependence. In simulations, we set $\Delta_j = \sigma$ for affected dimensions, activating the shift only when the lagged state condition holds.

Together, these anomaly types span persistent and transient deviations, first- and second-order structural changes, and local versus cross-sectional disruptions. This breadth prevents the evaluation from favoring a single modeling paradigm and instead forces methods to adapt across fundamentally different instability mechanisms. Performance is evaluated using Precision, Recall, F1-score, False Positive Rate (FPR), and AUROC. In addition to the anomaly-specific results reported in Tables~\ref{tab:multi_metric_low_all_synthetic} and \ref{tab:multi_metric_high_all_synthetic}, Table~\ref{tab:synthetic_structural_summary} reports overall average performance across all anomaly types and contamination regimes. Models are ranked by average F1-score, providing a direct summary comparison while preserving visibility into regime-specific strengths and weaknesses.

The aggregate results indicate that ReGEN-TAD achieves the highest overall F1-score (0.6239), outperforming the next-best competitor, GARCH (0.5945), by a non-trivial margin. This improvement is accompanied by the highest overall precision (0.8340) and strong AUROC (0.9477), while maintaining a low average FPR (0.0092). Although GARCH attains slightly higher average recall (0.6261 versus 0.5954), this comes with materially lower precision and AUROC, indicating weaker ranking quality and less balanced discrimination. In contrast, ReGEN-TAD achieves the strongest joint performance profile across F1, precision, and AUROC, suggesting that its gains are not driven by aggressive thresholding but by improved structural separation between normal and anomalous regimes.

The anomaly-specific tables clarify where these aggregate gains originate. Under low contamination, ReGEN-TAD consistently attains the strongest or near-strongest F1-scores for mean, trend, and collective shifts, with recall typically exceeding 0.69 and FPR effectively near zero (often below 0.002). These regimes correspond to persistent structural deviations, where latent representation stability and ensemble aggregation provide clear advantages. Competing methods frequently exhibit lower recall for these structured shifts or require higher false positive rates to achieve comparable sensitivity. Variance and contextual shifts produce a different ranking pattern. Linear econometric baselines such as OLS and RRR achieve very high recall in variance regimes (often above 0.90 under low contamination) and correspondingly strong F1-scores, reflecting their direct sensitivity to dispersion changes. However, their performance deteriorates for persistent mean and trend shifts, where cross-sectional aggregation and latent modeling become more informative. This differentiation reinforces that no single detection principle dominates across all structural perturbations.

Spike anomalies remain the primary regime in which residual-based and forecasting-driven methods are comparatively strong. Because spikes induce immediate prediction failures, simple residual magnitude measures respond sharply. ReGEN-TAD achieves moderate yet stable performance in this setting (F1 approximately 0.56 under both contamination regimes), but its comparative advantage lies in persistent and collective deviations rather than purely transient shocks.

\begin{table}[H]
\centering
\scriptsize
\caption{Overall Average Performance Metrics across all Anomaly Types and Contamination Rates from Synthetic Data. Models are ranked by F1 Score.}
\label{tab:synthetic_structural_summary}
\begin{tabular}{lcccccc}
\toprule
Model & F1 & Precision & Recall & AUROC & FPR & Runtime (s) \\
\midrule
ReGEN-TAD & 0.6239 & 0.8340 & 0.5954 & 0.9477 & 0.0092 & 24.5519 \\
GARCH & 0.5945 & 0.6635 & 0.6261 & 0.8432 & 0.0228 & 0.0152 \\
OLS & 0.5457 & 0.6978 & 0.5175 & 0.9549 & 0.0042 & 3.0716 \\
RRR & 0.5141 & 0.6519 & 0.4931 & 0.8860 & 0.0038 & 3.6446 \\
TimeGPT & 0.4949 & 0.7437 & 0.4039 & 0.6970 & 0.0099 & 3.6237 \\
TranAD & 0.4395 & 0.6054 & 0.4614 & 0.7218 & 0.0148 & 8.0562 \\
DAGMM & 0.3525 & 0.5100 & 0.3406 & 0.7540 & 0.0149 & 1.9772 \\
TGAN-AD & 0.3261 & 0.4416 & 0.3214 & 0.7108 & 0.0170 & 16.7449 \\
DeepAnt & 0.3072 & 0.4909 & 0.2774 & 0.8161 & 0.0010 & 3.7376 \\
AlioghliOkay & 0.2925 & 0.4690 & 0.2723 & 0.8130 & 0.0010 & 7.6047 \\
IForest & 0.2734 & 0.3234 & 0.2929 & 0.7762 & 0.0142 & 0.1334 \\
\bottomrule
\end{tabular}
\end{table}

At higher contamination levels, signal strength increases for all methods, yet relative ordering remains broadly stable. ReGEN-TAD continues to deliver F1-scores above 0.83 for mean, trend, and contextual shifts while maintaining FPR at or near zero. Several competing methods exhibit substantial declines in recall under heavy contamination for trend and collective anomalies, indicating sensitivity to calibration instability or dilution effects in high-dimensional settings.

From an overall perspective, ReGEN-TAD displays the most stable performance profile across anomaly types. Its recall remains concentrated in the 0.59--0.71 range across persistent structural classes, whereas competing methods exhibit substantially greater dispersion, with recall varying from near zero to above 0.90 depending on anomaly type. Simultaneously, false alarm control remains consistently strong: the average FPR of 0.0092 is among the lowest of the leading methods, and improvements in recall do not translate into disproportionate increases in false positives. Taken together, the overall ranking and anomaly-specific breakdown demonstrate that ReGEN-TAD does not merely specialize in a narrow regime. While certain competitors dominate in isolated settings (e.g., variance or spike detection), their performance deteriorates outside those niches. In contrast, the proposed framework maintains consistently strong F1, high precision, robust AUROC, and controlled FPR across persistent, cross-sectional, and context-dependent structural deviations, supporting its robustness in heterogeneous high-dimensional environments.

\begin{sidewaystable}[p]
    \centering
    \scriptsize 
    
    \caption{Average performance metrics at low contamination levels ($\gamma \in \{0.01, 0.03, 0.05\}$). Models are ordered by overall F1 score (highest to lowest). Metrics are computed from integer window counts; repeated values reflect discrete partitioning and fixed quantile calibration.}
    \label{tab:multi_metric_low_all_synthetic}
    \begin{tabular}{lllccccccccccl}
    \toprule
    Shift Type & $(T,p)$ & Metric & ReGEN-TAD & GARCH & OLS & RRR & TimeGPT & TranAD & DAGMM & TGANAD & DeepAnt & AlioghliOkay & IForest \\
    \midrule
    Mean Shift & (500,100) & F1 & 0.8306 & 0.7406 & 0.5441 & 0.6284 & 0.3629 & 0.6512 & 0.3811 & 0.1088 & 0.3070 & 0.4362 & 0.5290 \\
     & & FPR & 0.0002 & 0.0278 & 0.0087 & 0.0083 & 0.0273 & 0.0269 & 0.0290 & 0.0394 & 0.0000 & 0.0044 & 0.0280 \\
     & & Precision & 0.9967 & 0.7038 & 0.6211 & 0.6806 & 0.4606 & 0.6455 & 0.4207 & 0.1167 & 0.5333 & 0.5122 & 0.5337 \\
     & & Recall & 0.7119 & 0.7976 & 0.5167 & 0.6190 & 0.3000 & 0.7024 & 0.3619 & 0.1048 & 0.2524 & 0.4143 & 0.5452 \\
    \midrule
    Trend Shift & (500,100)  & F1 & 0.8333 & 0.6456 & 0.3401 & 0.3385 & 0.2305 & 0.3105 & 0.0390 & 0.0678 & 0.1985 & 0.0222 & 0.0819 \\
     & & FPR & 0.0000 & 0.0371 & 0.0083 & 0.0085 & 0.0330 & 0.0269 & 0.0290 & 0.0394 & 0.0000 & 0.0004 & 0.0280 \\
     & & Precision & 1.0000 & 0.6260 & 0.5771 & 0.5450 & 0.3072 & 0.4492 & 0.0540 & 0.0865 & 0.4667 & 0.0267 & 0.1259 \\
     & & Recall & 0.7143 & 0.6929 & 0.2548 & 0.2524 & 0.1857 & 0.2500 & 0.0310 & 0.0571 & 0.1286 & 0.0190 & 0.0667 \\
    \midrule
    Variance Shift & (500,150)  & F1 & 0.8333 & 0.4165 & 0.8601 & 0.8588 & 0.7761 & 0.8587 & 0.5855 & 0.3242 & 0.2922 & 0.7326 & 0.6661 \\
     & & FPR & 0.0000 & 0.0513 & 0.0091 & 0.0091 & 0.0034 & 0.0269 & 0.0290 & 0.0394 & 0.0000 & 0.0110 & 0.0280 \\
     & & Precision & 1.0000 & 0.4222 & 0.8322 & 0.8324 & 0.9333 & 0.7594 & 0.6027 & 0.3626 & 0.5000 & 0.7512 & 0.6711 \\
     & & Recall & 0.7143 & 0.4452 & 0.9048 & 0.9048 & 0.6643 & 0.9976 & 0.6000 & 0.3048 & 0.2143 & 0.7786 & 0.6786 \\
    \midrule
    Spike & (500,100)  & F1 & 0.5583 & 0.1141 & 0.6465 & 0.6283 & 0.2184 & 0.1712 & 0.2129 & 0.1965 & 0.6115 & 0.5361 & 0.1433 \\
     & & FPR & 0.0187 & 0.0434 & 0.0083 & 0.0087 & 0.0341 & 0.0269 & 0.0290 & 0.0394 & 0.0133 & 0.0000 & 0.0280 \\
     & & Precision & 0.6700 & 0.3551 & 0.8391 & 0.8306 & 0.2826 & 0.2674 & 0.2773 & 0.2253 & 0.8549 & 1.0000 & 0.2144 \\
     & & Recall & 0.4786 & 0.0929 & 0.5643 & 0.5405 & 0.1786 & 0.1357 & 0.1833 & 0.1786 & 0.5238 & 0.3738 & 0.1119 \\
    \midrule
    Collective & (500,100)  & F1 & 0.8056 & 0.7877 & 0.5927 & 0.6066 & 0.2935 & 0.4487 & 0.3334 & 0.0957 & 0.0703 & 0.4189 & 0.4534 \\
     & & FPR & 0.0019 & 0.0295 & 0.0087 & 0.0083 & 0.0324 & 0.0269 & 0.0290 & 0.0394 & 0.0057 & 0.0028 & 0.0280 \\
     & & Precision & 0.9667 & 0.7166 & 0.6260 & 0.6181 & 0.3556 & 0.4055 & 0.3830 & 0.1089 & 0.0590 & 0.4364 & 0.4677 \\
     & & Recall & 0.6905 & 0.8929 & 0.6024 & 0.6286 & 0.2500 & 0.5143 & 0.3119 & 0.0881 & 0.0881 & 0.4238 & 0.4714 \\
    \midrule
    Contextual & (500,100) & F1 & 0.8333 & 0.7461 & 0.8769 & 0.8653 & 0.7362 & 0.8575 & 0.4451 & 0.2729 & 0.3598 & 0.4893 & 0.6101 \\
     & & FPR & 0.0000 & 0.0366 & 0.0087 & 0.0083 & 0.0057 & 0.0269 & 0.0290 & 0.0394 & 0.0000 & 0.0034 & 0.0280 \\
     & & Precision & 1.0000 & 0.6779 & 0.8367 & 0.8408 & 0.8889 & 0.7587 & 0.5249 & 0.2909 & 0.5000 & 0.4967 & 0.6403 \\
     & & Recall & 0.7143 & 0.8571 & 0.9238 & 0.9048 & 0.6286 & 0.9952 & 0.4143 & 0.2690 & 0.3143 & 0.4976 & 0.6048 \\
    \bottomrule
    \end{tabular}

    \vspace{2em} 

    \caption{Average performance metrics at high contamination levels ($\gamma \in \{0.10, 0.12, 0.15\}$). Models ordered by overall F1 score (highest to lowest).}
    \label{tab:multi_metric_high_all_synthetic}
    \begin{tabular}{lllccccccccccl}
    \toprule
    Shift Type & $(T,p)$ & Metric & ReGEN-TAD & GARCH & OLS & RRR & TimeGPT & TranAD & DAGMM & TGANAD & DeepAnt & AlioghliOkay & IForest \\
    \midrule
    Mean Shift & (500,100)  & F1 & 0.8333 & 0.7504 & 0.2996 & 0.3496 & 0.3629 & 0.4203 & 0.0948 & 0.0600 & 0.0890 & 0.0619 & 0.0044 \\
     & & FPR & 0.0000 & 0.0119 & 0.0025 & 0.0021 & 0.0273 & 0.0000 & 0.0004 & 0.0104 & 0.0002 & 0.0000 & 0.0000 \\
     & & Precision & 1.0000 & 0.8331 & 0.4377 & 0.4576 & 0.4606 & 0.7667 & 0.3278 & 0.1137 & 0.0978 & 0.1000 & 0.0333 \\
     & & Recall & 0.7143 & 0.7262 & 0.2500 & 0.3095 & 0.3000 & 0.3500 & 0.0595 & 0.0429 & 0.0833 & 0.0500 & 0.0024 \\
    \midrule
    Trend Shift & (500,100)  & F1 & 0.8333 & 0.5859 & 0.0893 & 0.0703 & 0.2305 & 0.0326 & 0.0000 & 0.0460 & 0.0383 & 0.0000 & 0.0000 \\
     & & FPR & 0.0000 & 0.0375 & 0.0011 & 0.0009 & 0.0330 & 0.0000 & 0.0004 & 0.0104 & 0.0000 & 0.0000 & 0.0000 \\
     & & Precision & 1.0000 & 0.7117 & 0.2072 & 0.1271 & 0.3072 & 0.1667 & 0.0000 & 0.1022 & 0.1000 & 0.0000 & 0.0000 \\
     & & Recall & 0.7143 & 0.5714 & 0.0619 & 0.0500 & 0.1857 & 0.0190 & 0.0000 & 0.0310 & 0.0238 & 0.0000 & 0.0000 \\
    \midrule
    Variance Shift & (500,100)  & F1 & 0.8333 & 0.3082 & 0.7187 & 0.6661 & 0.7761 & 0.8120 & 0.3242 & 0.2809 & 0.0266 & 0.4927 & 0.0089 \\
     & & FPR & 0.0000 & 0.0456 & 0.0044 & 0.0036 & 0.0034 & 0.0000 & 0.0004 & 0.0104 & 0.0000 & 0.0042 & 0.0000 \\
     & & Precision & 1.0000 & 0.4107 & 0.7494 & 0.7252 & 0.9333 & 0.9333 & 0.6852 & 0.5148 & 0.0667 & 0.5205 & 0.0667 \\
     & & Recall & 0.7143 & 0.3048 & 0.7238 & 0.6619 & 0.6643 & 0.7714 & 0.2262 & 0.2143 & 0.0167 & 0.5048 & 0.0048 \\
    \midrule
    Spike & (500,100)  & F1 & 0.5583 & 0.0938 & 0.5720 & 0.5451 & 0.2184 & 0.0933 & 0.2108 & 0.1659 & 0.5544 & 0.5339 & 0.0000 \\
     & & FPR & 0.0187 & 0.0282 & 0.0006 & 0.0008 & 0.0341 & 0.0000 & 0.0004 & 0.0104 & 0.0008 & 0.0000 & 0.0000 \\
     & & Precision & 0.6700 & 0.4633 & 0.9825 & 0.9763 & 0.2826 & 0.7000 & 0.5333 & 0.3065 & 0.9832 & 1.0000 & 0.0000 \\
     & & Recall & 0.4786 & 0.0810 & 0.4238 & 0.3881 & 0.1786 & 0.0500 & 0.1452 & 0.1190 & 0.4024 & 0.3714 & 0.0000 \\
    \midrule
    Collective & (500,100)  & F1 & 0.7306 & 0.7966 & 0.2178 & 0.2886 & 0.2935 & 0.3333 & 0.0957 & 0.0332 & 0.0321 & 0.0417 & 0.0089 \\
     & & FPR & 0.0070 & 0.0127 & 0.0023 & 0.0021 & 0.0324 & 0.0000 & 0.0004 & 0.0104 & 0.0000 & 0.0000 & 0.0000 \\
     & & Precision & 0.8767 & 0.7946 & 0.3085 & 0.3136 & 0.3556 & 0.3333 & 0.2933 & 0.0645 & 0.0333 & 0.0667 & 0.0667 \\
     & & Recall & 0.6262 & 0.8238 & 0.1833 & 0.2929 & 0.2500 & 0.3333 & 0.0619 & 0.0238 & 0.0310 & 0.0381 & 0.0048 \\
    \midrule
    Contextual & (500,100)  & F1 & 0.8333 & 0.7715 & 0.5160 & 0.4194 & 0.7362 & 0.5568 & 0.1504 & 0.1630 & 0.0000 & 0.0442 & 0.0000 \\
     & & FPR & 0.0000 & 0.0205 & 0.0027 & 0.0017 & 0.0057 & 0.0000 & 0.0004 & 0.0087 & 0.0000 & 0.0000 & 0.0000 \\
     & & Precision & 1.0000 & 0.7976 & 0.5361 & 0.4803 & 0.8889 & 0.6667 & 0.5389 & 0.3777 & 0.0000 & 0.0667 & 0.0000 \\
     & & Recall & 0.7143 & 0.7976 & 0.5143 & 0.4071 & 0.6286 & 0.5095 & 0.0952 & 0.1262 & 0.0000 & 0.0333 & 0.0000 \\
    \bottomrule
    \end{tabular}

\end{sidewaystable}

\subsection{Market Regime and Cross-Sectional Shock Simulations}

We evaluate detection performance in financially structured simulations, with results reported in Tables~\ref{tab:stock_all_metrics_transposed_stock_low}, \ref{tab:stock_all_metrics_transposed_stock_high}, and the aggregate ranking in Table~\ref{tab:stock_overall_summary_updated}. The first two tables present anomaly-specific results under low and high contamination, while the third reports averages across anomaly types and contamination levels, providing a unified F1-based ranking.

Together, these tables assess three dimensions: (i) stability across economically distinct anomaly classes, (ii) robustness to contamination in the calibration window, and (iii) aggregate performance across heterogeneous disturbances. Separating low- and high-contamination environments clarifies whether methods depend on stable calibration or remain reliable when training data are partially contaminated. The overall summary prevents rankings from being driven by isolated anomaly categories. The evaluation design spans 3,900 independent Monte Carlo simulation runs across anomaly classes, contamination regimes, and rolling window specifications. Again, this breadth of experimentation ensures that reported rankings reflect systematic performance differences rather than isolated  realizations.

\vspace{0.5em}
\noindent\textbf{Market direction shifts (Bear and Bull).}
Market-wide directional regimes correspond to sustained displacement in expected returns across a subset $\mathcal{J} \subset \{1,\dots,p\}$ of assets. A stylized formulation is
\begin{equation}
\mathbf{R}_{t+k}
=
\boldsymbol{\mu}
+
\boldsymbol{\Delta}
+
\boldsymbol{\varepsilon}_{t+k},
\qquad 
\boldsymbol{\Delta}_j
=
\begin{cases}
\delta, & j \in \mathcal{J}, \\
0, & j \notin \mathcal{J},
\end{cases}
\end{equation}
where $\delta > 0$ represents a bull regime and $\delta < 0$ a bear regime. These shifts are persistent and cross-sectional, altering the latent center of the return distribution. Forecasting models may gradually adapt as the new mean becomes incorporated into recent observations, whereas latent-embedding methods detect the sustained displacement relative to the calibrated baseline. In simulations, we set $|\mathcal{J}| = 0.5p$ and $\delta = \pm 0.02$.

\vspace{0.5em}
\noindent\textbf{Volatility spikes and liquidity dry-up.}
Volatility and liquidity disruptions primarily alter second-order structure without deterministic drift:
\begin{equation}
\mathbf{R}_{t+k}
\sim
\mathcal{N}(\boldsymbol{\mu},\, \boldsymbol{\Sigma}_1),
\qquad
\boldsymbol{\Sigma}_1 \succ \boldsymbol{\Sigma}_0,
\end{equation}
where $\boldsymbol{\Sigma}_1$ exhibits inflated diagonal entries (volatility spike) and potentially increased cross-sectional correlation (liquidity stress). Forecast residual magnitudes increase, but mean predictions remain approximately unbiased. In simulations, we double the diagonal variances and increase pairwise correlations within affected assets to $0.6$.

\vspace{0.5em}
\noindent\textbf{Regime switches.}
A regime switch combines mean and variance displacement:
\begin{equation}
\mathbf{R}_{t+k}
=
\boldsymbol{\mu}_1
+
\boldsymbol{\varepsilon}_{t+k},
\qquad
\boldsymbol{\varepsilon}_{t+k}
\sim
\mathcal{N}(\mathbf{0}, \boldsymbol{\Sigma}_1),
\end{equation}
with $(\boldsymbol{\mu}_1, \boldsymbol{\Sigma}_1) \neq (\boldsymbol{\mu}_0, \boldsymbol{\Sigma}_0)$. Such transitions emulate macroeconomic structural breaks. Because both location and dispersion shift simultaneously, these anomalies induce persistent embedding displacement and covariance distortion, favoring models that explicitly learn baseline latent geometry. In simulations, we set $\boldsymbol{\mu}_1 = \boldsymbol{\mu}_0 + 0.02\,\mathbf{1}$ and inflate variances by a factor of two.

\vspace{0.5em}
\noindent\textbf{Correlation breakdown.}
Correlation breakdown alters dependence structure while preserving marginal means:
\begin{equation}
\mathbf{R}_{t+k}
=
\boldsymbol{\mu}
+
\mathbf{B}_1 \mathbf{f}_{t+k}
+
\boldsymbol{\varepsilon}_{t+k},
\end{equation}
where $\mathbf{B}_1$ differs from the baseline loading matrix $\mathbf{B}_0$, thereby disrupting common factor structure. Although marginal variances may remain stable, covariance geometry changes. Detection therefore requires sensitivity to cross-sectional dependence rather than purely univariate deviations. In simulations, we perturb baseline loadings by adding Gaussian noise with standard deviation $0.3$ to affected rows of $\mathbf{B}_0$.

\vspace{0.5em}
\noindent\textbf{Contagion processes.}
Contagion represents gradual propagation of shocks across assets:
\begin{equation}
R_{t+k,j}
=
\mu_j
+
\delta \,\mathbb{I}(j \in \mathcal{J}_{k})
+
\varepsilon_{t+k,j},
\end{equation}
where $\mathcal{J}_{k}$ expands over time. Unlike simultaneous collective shifts, contagion introduces dynamic cross-sectional spread. Forecasting residuals increase sequentially, while latent embeddings exhibit expanding dispersion patterns across dimensions. In simulations, we initialize $|\mathcal{J}_{0}| = 0.1p$ and expand the set by $0.1p$ per step, using $\delta = 0.02$.

\vspace{0.5em}
\noindent\textbf{Momentum crashes and trend reversals.}
Momentum reversals correspond to sign changes in previously persistent drift:
\begin{equation}
R_{t+k,j}
=
\mu_j
+
\beta_j k \cdot \mathbb{I}(k < k^\star)
-
\beta_j (k-k^\star) \cdot \mathbb{I}(k \ge k^\star)
+
\varepsilon_{t+k,j}.
\end{equation}
These regimes create structured prediction failure as extrapolated trends abruptly invert. Longer forecast horizons amplify cumulative error, while generative embeddings reflect instability in temporal dynamics. In simulations, we set $\beta_j = 0.01$ for affected assets and $k^\star$ at the midpoint of the anomaly window.

\vspace{0.5em}
\noindent\textbf{Flash crashes and fat-tail events.}
Flash crashes correspond to isolated extreme realizations,
\begin{equation}
R_{t+k,j}
=
\mu_j
+
\delta \,\mathbb{I}(k = k^\star)
+
\varepsilon_{t+k,j},
\end{equation}
whereas fat-tail events arise from heavy-tailed innovations,
\begin{equation}
\varepsilon_{t+k,j} \sim t_\nu(0, \sigma_j^2).
\end{equation}
These events produce abrupt but often transient prediction failures. Because the detection statistic typically depends on the maximum discrepancy within the horizon window, performance saturates once the extreme deviation is included. In simulations, we set $\delta = -0.05$ and use $\nu = 3$ for heavy-tailed shocks.

\vspace{0.5em}
\noindent\textbf{Microstructure noise.}
Microstructure disturbances introduce high-frequency oscillatory components,
\begin{equation}
R_{t+k,j}
=
\mu_j
+
\eta_j \sin(\omega k)
+
\varepsilon_{t+k,j},
\end{equation}
creating localized bursts of instability without persistent regime displacement. In simulations, we set $\eta_j = 0.01$ for affected assets and $\omega = 2\pi/5$ to generate short-lived oscillatory perturbations.

Performance results are reported in Tables~\ref{tab:stock_all_metrics_transposed_stock_low} and~\ref{tab:stock_all_metrics_transposed_stock_high}, with aggregate averages in Table~\ref{tab:stock_overall_summary_updated}. The overall ranking shows that ReGEN-TAD attains the highest mean F1-score (0.7717), combining very high precision (0.9684) with stable recall (0.6427) and an extremely low false positive rate (0.0018). The gains are not driven by aggressive thresholding but by structural discrimination: across contamination regimes, precision remains near one in persistent and cross-sectional disturbances while recall stays consistently above 0.64.

Inspection of the anomaly-specific tables clarifies the source of these improvements. For persistent market-wide regimes (bear, bull, regime shifts, sector disturbances, and liquidity dry-ups), ReGEN-TAD delivers F1-scores clustered around 0.80 under both contamination levels, with FPR effectively zero in nearly all categories. This contrasts with methods whose performance varies sharply across regimes. Although OLS and RRR residual monitoring perform strongly in low-contamination covariance distortions, often exceeding 0.90 F1 for correlation breakdown, liquidity stress, and regime shifts, their performance deteriorates under higher contamination for directional shifts (e.g., bear/bull), where F1 falls below 0.31 for OLS and 0.29 for RRR. This pattern suggests reliance on stable residual structure rather than adaptive latent geometry.

Forecasting-based and transformer architectures display more heterogeneous behavior. TranAD achieves very high recall in several high-contamination structural regimes (often above 0.93), but with lower precision in directional shifts and weaker results in flash or microstructure settings. TimeGPT maintains comparatively strong precision (0.7795 overall) but lower recall (0.4348 overall), indicating that while pretrained forecasting models capture normal drift effectively, they are less consistent when deviations are structurally persistent yet subtle in marginal projections.

Volatility-focused modeling shows a complementary pattern. GARCH performs competitively in volatility spike and directional regimes under high contamination (F1 exceeding 0.84 for bear/bull markets), reflecting sensitivity to conditional variance amplification. However, performance declines in correlation breakdown and microstructure disturbances, where cross-sectional structure shifts without proportional variance escalation. The absence of joint latent modeling limits robustness in contagion and geometry-driven disruptions. Across contamination levels, ReGEN-TAD's F1 values remain tightly concentrated relative to competitors, whose dispersion is considerably larger. Competing methods dominate in isolated niches, such as linear residual models in stable covariance distortions, GARCH in variance surges, or transformer-based forecasting under abrupt shocks, but degrade outside those conditions. In contrast, the proposed ensemble maintains near-constant precision, controlled FPR, and stable recall across persistent, cross-sectional, and temporally structured disturbances.

\begin{table}[H]
\centering
\scriptsize
\caption{Overall Average Performance Metrics across all Anomaly Types and Contamination Rates. Models are ranked by F1 Score.}
\label{tab:stock_overall_summary_updated}
\begin{tabular}{lcccccc}
\toprule
Model & F1 & Precision & Recall & AUROC & FPR & Runtime (s) \\
\midrule
ReGEN-TAD & 0.7717 & 0.9684 & 0.6427 & 0.9704 & 0.0018 & 12.3828 \\
OLS & 0.6638 & 0.7636 & 0.6499 & 0.9747 & 0.0044 & 4.4457 \\
RRR & 0.6165 & 0.7197 & 0.6048 & 0.9364 & 0.0040 & 5.0803 \\
TranAD & 0.5612 & 0.7010 & 0.5721 & 0.7977 & 0.0146 & 4.4906 \\
TimeGPT & 0.5307 & 0.7795 & 0.4348 & 0.7132 & 0.0084 & 3.8887 \\
GARCH & 0.4905 & 0.5690 & 0.4874 & 0.7840 & 0.0292 & 0.0117 \\
AlioghliOkay & 0.3755 & 0.4921 & 0.3623 & 0.8766 & 0.0023 & 3.4801 \\
DAGMM & 0.3589 & 0.4959 & 0.3340 & 0.7697 & 0.0151 & 1.1285 \\
DeepAnt & 0.3298 & 0.4755 & 0.2957 & 0.8824 & 0.0009 & 2.2501 \\
IForest & 0.2916 & 0.3460 & 0.2923 & 0.8028 & 0.0143 & 0.1250 \\
TGAN-AD & 0.2878 & 0.3856 & 0.2671 & 0.6694 & 0.0183 & 8.3257 \\
\bottomrule
\end{tabular}
\end{table}

\begin{figure}[t]
    \centering
    \includegraphics[width=0.4\textwidth]{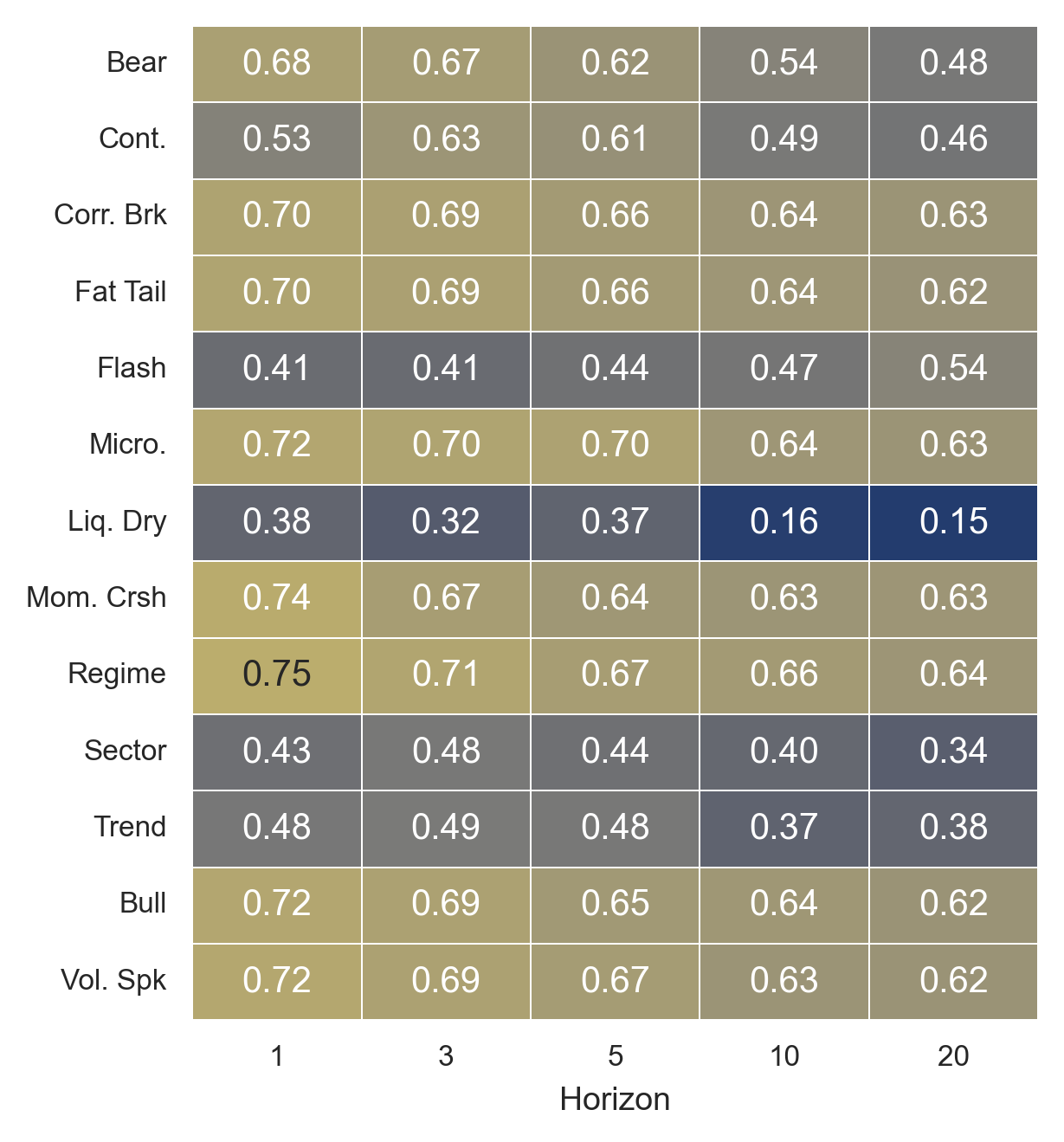}
    \caption{F1 score across prediction horizons and anomaly types. 
    Short- to medium-term horizons yield stronger overall detection performance.}
    \label{fig:horizon_heatmap}
\end{figure}

\subsection{Sensitivity to Prediction Horizon}

\begin{figure*}[t]
    \centering
    \includegraphics[width=\textwidth]{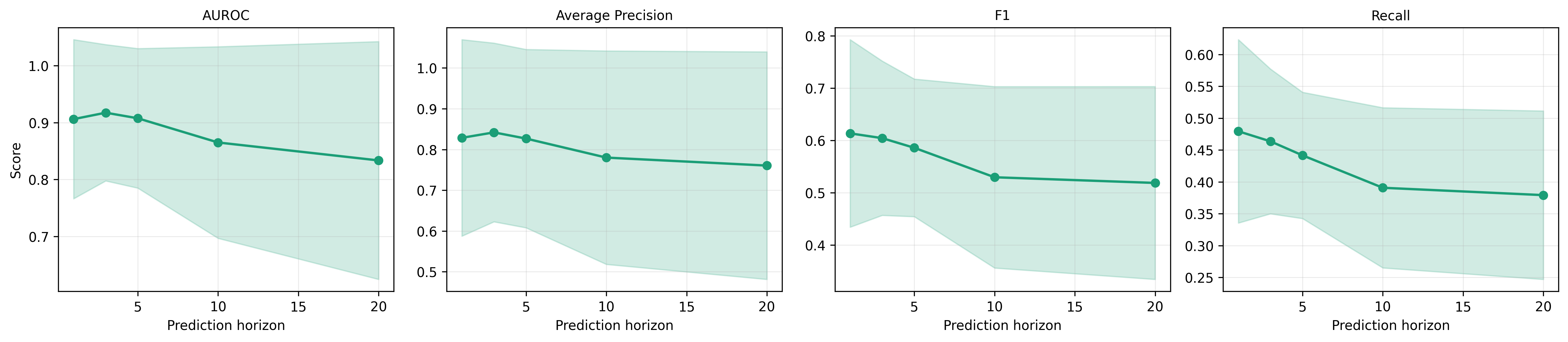}
    \caption{\textbf{ReGEN-TAD performance across prediction horizons.}
    Aggregate detection metrics (AUROC, Average Precision, F1, and Recall) 
    are reported for $H \in \{1,3,5,10,20\}$. Ranking-based measures 
    (AUROC and AP) remain comparatively stable across horizons, whereas 
    threshold-sensitive metrics (F1 and Recall) exhibit gradual attenuation 
    as forecast uncertainty increases with longer prediction horizons.}
    \label{fig:horizon_performance}
\end{figure*}

We evaluate ReGEN-TAD across horizons $H \in \{1,3,5,10,20\}$, spanning short- to medium-term forecasting scales. Figure~\ref{fig:horizon_performance} (Figure 8) reports aggregate AUROC, Average Precision, F1, and Recall across horizons, while Figure~\ref{fig:horizon_heatmap} provides anomaly-specific F1 behavior. As shown in Figure~\ref{fig:horizon_performance}, overall F1 and Recall are strongest at short horizons and decline gradually as $H$ increases. In contrast, AUROC remains comparatively stable, indicating that ranking quality is preserved even as threshold-based classification weakens. This divergence reflects the accumulation of forecast uncertainty in multi-step prediction, which elevates baseline error variance and reduces separation between normal and anomalous regimes.

The heatmap in Figure~\ref{fig:horizon_heatmap} reveals meaningful heterogeneity across anomaly types. Persistent structural disruptions, including regime shifts, volatility spikes, and correlation breakdowns, remain comparatively stable across horizons, with only moderate degradation as $H$ increases. In contrast, liquidity dry-ups and directional shifts exhibit sharper declines at longer horizons, indicating greater sensitivity to calibration noise and horizon-induced smoothing. Flash events show mild improvement at longer horizons, consistent with the idea that short-lived shocks become more detectable once their immediate propagation effects are incorporated into the forecast window.

Together, Figures~\ref{fig:horizon_performance} and~\ref{fig:horizon_heatmap} highlight a structural trade-off inherent in forecast-driven detection. Short horizons emphasize immediate prediction breakdowns and maximize sensitivity to abrupt deviations. Longer horizons integrate broader temporal information and may better reflect secondary market adjustments, but they also amplify forecast dispersion, raising the effective noise floor of the generative backbone. In practice, moderate horizons appear to balance responsiveness and stability, preserving detection power for persistent regimes while limiting degradation associated with long-horizon uncertainty.


    \clearpage
\begin{table}[H]
\centering
\scriptsize
\caption{Performance metrics by Model and Anomaly Type (Low Contamination $\gamma \in \{0.01, 0.03, 0.05\}$). Models are ordered by overall F1 score.}
\label{tab:stock_all_metrics_transposed_stock_low}
\resizebox{\textwidth}{!}{
\begin{tabular}{llccccccccccccc}
\toprule
Model & Metric & Bear & Bull & Cont. & Corr. Brk & Fat Tail & Flash & Liq. Dry & Micro. & Mom. Crsh & Regime & Sector & Trend & Vol. Spk \\
\midrule
ReGEN-TAD & F1 & 0.7831 & 0.7831 & 0.7944 & 0.8056 & 0.8056 & 0.5287 & 0.8056 & 0.7843 & 0.8056 & 0.8056 & 0.8056 & 0.7859 & 0.7859 \\
 & FPR & 0.0002 & 0.0002 & 0.0008 & 0.0000 & 0.0000 & 0.0189 & 0.0000 & 0.0015 & 0.0000 & 0.0000 & 0.0000 & 0.0000 & 0.0000 \\
 & Precision & 0.9967 & 0.9967 & 0.9862 & 1.0000 & 1.0000 & 0.6723 & 1.0000 & 0.9739 & 1.0000 & 1.0000 & 1.0000 & 1.0000 & 1.0000 \\
 & Recall & 0.6464 & 0.6464 & 0.6660 & 0.6753 & 0.6753 & 0.4367 & 0.6753 & 0.6573 & 0.6753 & 0.6753 & 0.6753 & 0.6488 & 0.6488 \\
 & Runtime (s) & 13.9892 & 14.0268 & 10.5484 & 10.5476 & 10.6600 & 14.3892 & 17.3604 & 10.4007 & 10.6488 & 10.4999 & 10.4864 & 14.2313 & 14.3094 \\
\midrule
OLS & F1 & 0.6901 & 0.6901 & 0.8955 & 0.9085 & 0.9204 & 0.5073 & 0.9477 & 0.5478 & 0.9468 & 0.9419 & 0.8983 & 0.5059 & 0.9180 \\
 & FPR & 0.0071 & 0.0071 & 0.0068 & 0.0071 & 0.0072 & 0.0067 & 0.0076 & 0.0067 & 0.0077 & 0.0079 & 0.0074 & 0.0067 & 0.0075 \\
 & Precision & 0.7918 & 0.7918 & 0.8749 & 0.8676 & 0.9036 & 0.8126 & 0.9036 & 0.7500 & 0.9036 & 0.9009 & 0.8643 & 0.7596 & 0.8782 \\
 & Recall & 0.6788 & 0.6788 & 0.9297 & 0.9571 & 0.9551 & 0.4041 & 0.9995 & 0.4531 & 0.9981 & 0.9914 & 0.9437 & 0.3975 & 0.9676 \\
 & Runtime (s) & 4.5229 & 4.4314 & 4.4337 & 4.4555 & 4.4463 & 4.4254 & 4.4273 & 4.4259 & 4.4762 & 4.4271 & 4.4375 & 4.4434 & 4.4049 \\
\midrule
RRR & F1 & 0.7539 & 0.7539 & 0.8966 & 0.8985 & 0.9028 & 0.4815 & 0.9423 & 0.5137 & 0.9485 & 0.9422 & 0.9105 & 0.4822 & 0.9198 \\
 & FPR & 0.0066 & 0.0066 & 0.0064 & 0.0065 & 0.0066 & 0.0066 & 0.0073 & 0.0065 & 0.0074 & 0.0079 & 0.0070 & 0.0065 & 0.0071 \\
 & Precision & 0.8358 & 0.8358 & 0.8733 & 0.8667 & 0.9021 & 0.7966 & 0.9008 & 0.7502 & 0.9068 & 0.9017 & 0.8699 & 0.7506 & 0.8821 \\
 & Recall & 0.7502 & 0.7502 & 0.9286 & 0.9402 & 0.9270 & 0.3808 & 0.9914 & 0.4149 & 0.9981 & 0.9910 & 0.9585 & 0.3711 & 0.9676 \\
 & Runtime (s) & 5.2217 & 5.1450 & 5.1019 & 5.0876 & 5.0823 & 5.0829 & 5.1065 & 5.0321 & 5.0670 & 5.0265 & 5.0779 & 5.1091 & 5.0493 \\
\midrule
TranAD & F1 & 0.6606 & 0.6636 & 0.7886 & 0.8613 & 0.8416 & 0.1264 & 0.8625 & 0.1862 & 0.8624 & 0.8622 & 0.8612 & 0.3147 & 0.8643 \\
 & FPR & 0.0289 & 0.0290 & 0.0282 & 0.0281 & 0.0281 & 0.0290 & 0.0281 & 0.0282 & 0.0282 & 0.0282 & 0.0283 & 0.0288 & 0.0287 \\
 & Precision & 0.6575 & 0.6605 & 0.7334 & 0.7624 & 0.7555 & 0.2227 & 0.7624 & 0.3006 & 0.7623 & 0.7620 & 0.7614 & 0.4493 & 0.7646 \\
 & Recall & 0.7011 & 0.7040 & 0.8709 & 0.9973 & 0.9601 & 0.0932 & 1.0000 & 0.1442 & 1.0000 & 1.0000 & 0.9988 & 0.2509 & 1.0000 \\
 & Runtime (s) & 5.1497 & 5.1173 & 3.8996 & 3.8737 & 3.9488 & 5.2818 & 3.8304 & 3.9664 & 3.9210 & 3.8380 & 3.8268 & 5.3147 & 5.2803 \\
\midrule
TimeGPT & F1 & 0.4816 & 0.4816 & 0.3909 & 0.6505 & 0.5046 & 0.2440 & 0.7108 & 0.4120 & 0.6297 & 0.7101 & 0.5228 & 0.4617 & 0.6981 \\
 & FPR & 0.0102 & 0.0102 & 0.0139 & 0.0042 & 0.0093 & 0.0217 & 0.0011 & 0.0130 & 0.0038 & 0.0011 & 0.0082 & 0.0109 & 0.0018 \\
 & Precision & 0.7399 & 0.7399 & 0.6333 & 0.8981 & 0.7618 & 0.4348 & 0.9617 & 0.6599 & 0.8938 & 0.9609 & 0.7841 & 0.7182 & 0.9472 \\
 & Recall & 0.3776 & 0.3776 & 0.2978 & 0.5599 & 0.4014 & 0.1813 & 0.6190 & 0.3157 & 0.5211 & 0.6185 & 0.4153 & 0.3597 & 0.6076 \\
 & Runtime (s) & 3.8294 & 3.9798 & 3.5660 & 4.0714 & 3.7888 & 4.4156 & 3.8831 & 3.9070 & 3.7717 & 4.0997 & 3.7906 & 3.8022 & 3.7850 \\
\midrule
GARCH & F1 & 0.7679 & 0.7679 & 0.7480 & 0.2341 & 0.2436 & 0.0783 & 0.4075 & 0.0798 & 0.1487 & 0.7544 & 0.8222 & 0.6631 & 0.4148 \\
 & FPR & 0.0294 & 0.0294 & 0.0362 & 0.0495 & 0.0499 & 0.0334 & 0.0453 & 0.0514 & 0.0508 & 0.0244 & 0.0264 & 0.0343 & 0.0441 \\
 & Precision & 0.7234 & 0.7234 & 0.6758 & 0.2739 & 0.2862 & 0.2827 & 0.4383 & 0.1022 & 0.1849 & 0.7532 & 0.7530 & 0.6573 & 0.4474 \\
 & Recall & 0.8339 & 0.8339 & 0.8576 & 0.2162 & 0.2216 & 0.0581 & 0.3978 & 0.0751 & 0.1355 & 0.7773 & 0.9192 & 0.6865 & 0.4032 \\
 & Runtime (s) & 0.0134 & 0.0124 & 0.0128 & 0.0122 & 0.0129 & 0.0127 & 0.0118 & 0.0126 & 0.0128 & 0.0126 & 0.0114 & 0.0127 & 0.0127 \\
\midrule
AlioghliOkay & F1 & 0.4107 & 0.4105 & 0.3386 & 0.2135 & 0.4064 & 0.2930 & 0.8483 & 0.0190 & 0.9326 & 0.9359 & 0.6741 & 0.0192 & 0.8261 \\
 & FPR & 0.0020 & 0.0020 & 0.0002 & 0.0011 & 0.0014 & 0.0003 & 0.0078 & 0.0001 & 0.0099 & 0.0115 & 0.0055 & 0.0001 & 0.0060 \\
 & Precision & 0.5407 & 0.5407 & 0.5979 & 0.3379 & 0.6826 & 0.9451 & 0.8517 & 0.0476 & 0.8982 & 0.8822 & 0.7087 & 0.1542 & 0.8377 \\
 & Recall & 0.3724 & 0.3722 & 0.2683 & 0.1998 & 0.3407 & 0.1892 & 0.8843 & 0.0135 & 0.9853 & 0.9988 & 0.6956 & 0.0121 & 0.8510 \\
 & Runtime (s) & 4.1566 & 4.0534 & 2.8589 & 2.8650 & 2.9090 & 4.1273 & 4.6982 & 2.8729 & 2.9238 & 2.8408 & 2.7863 & 4.1370 & 4.1386 \\
\midrule
DAGMM & F1 & 0.4815 & 0.4815 & 0.3095 & 0.4223 & 0.4549 & 0.2973 & 0.7296 & 0.0961 & 0.7134 & 0.7481 & 0.6128 & 0.0674 & 0.7182 \\
 & FPR & 0.0297 & 0.0297 & 0.0293 & 0.0293 & 0.0293 & 0.0299 & 0.0293 & 0.0293 & 0.0293 & 0.0293 & 0.0293 & 0.0297 & 0.0297 \\
 & Precision & 0.4900 & 0.4900 & 0.3956 & 0.5056 & 0.5285 & 0.3939 & 0.7030 & 0.1537 & 0.6906 & 0.7091 & 0.6232 & 0.1151 & 0.6893 \\
 & Recall & 0.4950 & 0.4950 & 0.2682 & 0.3836 & 0.4222 & 0.2487 & 0.7759 & 0.0730 & 0.7650 & 0.8106 & 0.6303 & 0.0494 & 0.7723 \\
 & Runtime (s) & 1.2305 & 1.2230 & 1.0389 & 1.0364 & 1.0373 & 1.2500 & 1.0301 & 1.0317 & 1.0429 & 1.0247 & 1.0200 & 1.2656 & 1.2593 \\
\midrule
DeepAnt & F1 & 0.4709 & 0.4709 & 0.7976 & 0.5242 & 0.5876 & 0.3793 & 0.6171 & 0.2626 & 0.4359 & 0.7723 & 0.4435 & 0.2555 & 0.6693 \\
 & FPR & 0.0014 & 0.0014 & 0.0011 & 0.0013 & 0.0014 & 0.0053 & 0.0015 & 0.0013 & 0.0016 & 0.0019 & 0.0014 & 0.0001 & 0.0018 \\
 & Precision & 0.6894 & 0.6894 & 0.8379 & 0.6699 & 0.8509 & 0.8736 & 0.7181 & 0.4879 & 0.4837 & 0.8808 & 0.5184 & 0.5645 & 0.7367 \\
 & Recall & 0.4217 & 0.4217 & 0.7824 & 0.4834 & 0.5352 & 0.2890 & 0.5805 & 0.1902 & 0.4265 & 0.7335 & 0.4329 & 0.1737 & 0.6507 \\
 & Runtime (s) & 2.5258 & 2.5135 & 2.0119 & 2.0070 & 2.0308 & 2.6081 & 1.9813 & 1.9998 & 2.0120 & 1.9949 & 2.0094 & 2.5698 & 2.5606 \\
\midrule
IForest & F1 & 0.6180 & 0.6180 & 0.3959 & 0.6484 & 0.3754 & 0.1408 & 0.7689 & 0.0872 & 0.7776 & 0.7708 & 0.7489 & 0.1488 & 0.7752 \\
 & FPR & 0.0283 & 0.0283 & 0.0281 & 0.0281 & 0.0281 & 0.0286 & 0.0281 & 0.0281 & 0.0281 & 0.0281 & 0.0281 & 0.0283 & 0.0283 \\
 & Precision & 0.5991 & 0.5991 & 0.4667 & 0.6672 & 0.4788 & 0.2380 & 0.7264 & 0.1372 & 0.7294 & 0.7264 & 0.7141 & 0.2348 & 0.7307 \\
 & Recall & 0.6635 & 0.6635 & 0.3628 & 0.6529 & 0.3223 & 0.1050 & 0.8286 & 0.0681 & 0.8450 & 0.8326 & 0.8034 & 0.1152 & 0.8392 \\
 & Runtime (s) & 0.1263 & 0.1238 & 0.1188 & 0.1194 & 0.1229 & 0.1346 & 0.1193 & 0.1226 & 0.1198 & 0.1181 & 0.1241 & 0.1319 & 0.1339 \\
\midrule
TGAN-AD & F1 & 0.3701 & 0.3590 & 0.2545 & 0.2693 & 0.3204 & 0.2684 & 0.4776 & 0.0599 & 0.6016 & 0.5822 & 0.4450 & 0.0726 & 0.5954 \\
 & FPR & 0.0325 & 0.0321 & 0.0329 & 0.0328 & 0.0331 & 0.0327 & 0.0331 & 0.0327 & 0.0329 & 0.0329 & 0.0330 & 0.0324 & 0.0319 \\
 & Precision & 0.3807 & 0.3725 & 0.3028 & 0.3180 & 0.3603 & 0.3523 & 0.4813 & 0.1002 & 0.5765 & 0.5708 & 0.4388 & 0.1184 & 0.5795 \\
 & Recall & 0.3817 & 0.3704 & 0.2311 & 0.2430 & 0.2959 & 0.2253 & 0.4913 & 0.0445 & 0.6452 & 0.6181 & 0.4695 & 0.0547 & 0.6331 \\
 & Runtime (s) & 9.8110 & 9.7751 & 7.0184 & 7.0082 & 7.0462 & 10.2303 & 6.8738 & 7.0728 & 7.0213 & 6.8812 & 6.8397 & 10.1946 & 10.0682 \\
\bottomrule
\end{tabular}
}
\end{table}
\clearpage

\afterpage{
    \clearpage
\begin{table}[p] 
\centering
\scriptsize
\caption{Performance metrics by Model and Anomaly Type (High Contamination $\gamma \in \{0.10, 0.12, 0.15\}$). Models are ranked by overall F1 score.}
\label{tab:stock_all_metrics_transposed_stock_high}
\resizebox{\textwidth}{!}{
\begin{tabular}{llccccccccccccc}
\toprule
Model & Metric & Bear & Bull & Cont. & Corr. Brk & Fat Tail & Flash & Liq. Dry & Micro. & Mom. Crsh & Regime & Sector & Trend & Vol. Spk \\
\midrule
ReGEN-TAD & F1 & 0.7825 & 0.7825 & 0.7870 & 0.8056 & 0.8056 & 0.5194 & 0.8056 & 0.7750 & 0.8056 & 0.8056 & 0.8056 & 0.7816 & 0.7859 \\
 & FPR & 0.0002 & 0.0002 & 0.0013 & 0.0000 & 0.0000 & 0.0206 & 0.0000 & 0.0021 & 0.0000 & 0.0000 & 0.0000 & 0.0003 & 0.0000 \\
 & Precision & 0.9957 & 0.9957 & 0.9773 & 1.0000 & 1.0000 & 0.6438 & 1.0000 & 0.9627 & 1.0000 & 1.0000 & 1.0000 & 0.9947 & 1.0000 \\
 & Recall & 0.6460 & 0.6460 & 0.6597 & 0.6753 & 0.6753 & 0.4359 & 0.6753 & 0.6494 & 0.6753 & 0.6753 & 0.6753 & 0.6452 & 0.6488 \\
 & Runtime (s) & 13.9465 & 14.2630 & 10.5547 & 10.5617 & 10.4586 & 10.4556 & 10.5659 & 10.4706 & 10.6469 & 10.5352 & 10.4733 & 14.5496 & 14.2286 \\
\midrule
OLS & F1 & 0.3049 & 0.3049 & 0.4795 & 0.5087 & 0.5173 & 0.3928 & 0.9187 & 0.1594 & 0.8935 & 0.9537 & 0.5250 & 0.1573 & 0.8239 \\
 & FPR & 0.0014 & 0.0014 & 0.0007 & 0.0010 & 0.0011 & 0.0007 & 0.0024 & 0.0005 & 0.0033 & 0.0044 & 0.0021 & 0.0005 & 0.0025 \\
 & Precision & 0.4941 & 0.4941 & 0.5646 & 0.5813 & 0.7252 & 0.9459 & 0.9479 & 0.3662 & 0.8964 & 0.9512 & 0.6644 & 0.3830 & 0.8375 \\
 & Recall & 0.2637 & 0.2637 & 0.4506 & 0.4887 & 0.4688 & 0.2751 & 0.9145 & 0.1129 & 0.9050 & 0.9669 & 0.4971 & 0.1084 & 0.8278 \\
 & Runtime (s) & 4.4558 & 4.4796 & 4.4466 & 4.4411 & 4.4361 & 4.4661 & 4.4370 & 4.4214 & 4.4668 & 4.4397 & 4.4564 & 4.4455 & 4.4373 \\
\midrule
RRR & F1 & 0.2851 & 0.2851 & 0.2951 & 0.2276 & 0.2759 & 0.2682 & 0.8071 & 0.0489 & 0.8609 & 0.9455 & 0.5755 & 0.0462 & 0.7622 \\
 & FPR & 0.0008 & 0.0008 & 0.0003 & 0.0004 & 0.0004 & 0.0006 & 0.0018 & 0.0002 & 0.0027 & 0.0040 & 0.0013 & 0.0002 & 0.0021 \\
 & Precision & 0.4716 & 0.4716 & 0.4232 & 0.3482 & 0.5714 & 0.9437 & 0.8669 & 0.1369 & 0.8747 & 0.9554 & 0.6371 & 0.1504 & 0.7885 \\
 & Recall & 0.2456 & 0.2456 & 0.2583 & 0.2088 & 0.2143 & 0.1725 & 0.7944 & 0.0340 & 0.8682 & 0.9505 & 0.5704 & 0.0307 & 0.7610 \\
 & Runtime (s) & 5.1744 & 5.0330 & 5.0468 & 5.0528 & 5.0358 & 5.0882 & 5.0692 & 5.0345 & 5.0399 & 5.0793 & 5.1019 & 5.0515 & 5.0900 \\
\midrule
TranAD & F1 & 0.3427 & 0.3456 & 0.3062 & 0.3202 & 0.4234 & 0.0692 & 0.9550 & 0.1022 & 0.9538 & 0.9561 & 0.6341 & 0.0552 & 0.7970 \\
 & FPR & 0.0002 & 0.0002 & 0.0001 & 0.0002 & 0.0001 & 0.0002 & 0.0001 & 0.0002 & 0.0002 & 0.0001 & 0.0001 & 0.0002 & 0.0003 \\
 & Precision & 0.7490 & 0.7493 & 0.5145 & 0.5982 & 0.9819 & 0.7850 & 0.9986 & 0.6689 & 0.9982 & 0.9986 & 0.7319 & 0.4983 & 0.9086 \\
 & Recall & 0.2613 & 0.2652 & 0.2419 & 0.2750 & 0.3074 & 0.0367 & 0.9414 & 0.0561 & 0.9337 & 0.9351 & 0.6114 & 0.0309 & 0.7744 \\
 & Runtime (s) & 5.1553 & 5.2658 & 3.8980 & 3.8777 & 3.9678 & 3.8068 & 3.8561 & 3.9993 & 3.9370 & 3.8617 & 3.8427 & 5.3701 & 5.2901 \\
\midrule
TimeGPT & F1 & 0.4816 & 0.4816 & 0.3909 & 0.6505 & 0.5046 & 0.2440 & 0.7108 & 0.4120 & 0.6297 & 0.7101 & 0.5228 & 0.4617 & 0.6981 \\
 & FPR & 0.0102 & 0.0102 & 0.0139 & 0.0042 & 0.0093 & 0.0217 & 0.0011 & 0.0130 & 0.0038 & 0.0011 & 0.0082 & 0.0109 & 0.0018 \\
 & Precision & 0.7399 & 0.7399 & 0.6333 & 0.8981 & 0.7618 & 0.4348 & 0.9617 & 0.6599 & 0.8938 & 0.9609 & 0.7841 & 0.7182 & 0.9472 \\
 & Recall & 0.3776 & 0.3776 & 0.2978 & 0.5599 & 0.4014 & 0.1813 & 0.6190 & 0.3157 & 0.5211 & 0.6185 & 0.4153 & 0.3597 & 0.6076 \\
 & Runtime (s) & 3.7944 & 3.7897 & 3.5813 & 3.9293 & 3.9873 & 4.3042 & 3.9092 & 3.8496 & 3.8256 & 4.1000 & 3.7619 & 3.7915 & 3.7912 \\
\midrule
GARCH & F1 & 0.8469 & 0.8469 & 0.7700 & 0.1618 & 0.1742 & 0.0735 & 0.3387 & 0.0463 & 0.0865 & 0.8269 & 0.8971 & 0.6878 & 0.3481 \\
 & FPR & 0.0087 & 0.0087 & 0.0159 & 0.0337 & 0.0350 & 0.0170 & 0.0308 & 0.0357 & 0.0355 & 0.0059 & 0.0073 & 0.0174 & 0.0222 \\
 & Precision & 0.8937 & 0.8937 & 0.8150 & 0.3654 & 0.3565 & 0.5095 & 0.5758 & 0.1410 & 0.2036 & 0.9201 & 0.8966 & 0.8321 & 0.6098 \\
 & Recall & 0.8277 & 0.8277 & 0.7667 & 0.1369 & 0.1492 & 0.0552 & 0.2800 & 0.0559 & 0.0806 & 0.7685 & 0.9061 & 0.6183 & 0.2797 \\
 & Runtime (s) & 0.0103 & 0.0107 & 0.0113 & 0.0111 & 0.0111 & 0.0104 & 0.0109 & 0.0106 & 0.0110 & 0.0106 & 0.0103 & 0.0113 & 0.0106 \\
\midrule
AlioghliOkay & F1 & 0.0256 & 0.0256 & 0.0071 & 0.0000 & 0.0827 & 0.3813 & 0.5940 & 0.0000 & 0.8339 & 0.9438 & 0.2998 & 0.0000 & 0.6161 \\
 & FPR & 0.0000 & 0.0000 & 0.0000 & 0.0000 & 0.0000 & 0.0000 & 0.0018 & 0.0000 & 0.0041 & 0.0070 & 0.0011 & 0.0000 & 0.0025 \\
 & Precision & 0.1111 & 0.1111 & 0.0500 & 0.0000 & 0.2167 & 1.0000 & 0.6804 & 0.0000 & 0.9059 & 0.9240 & 0.4040 & 0.0000 & 0.6513 \\
 & Recall & 0.0188 & 0.0188 & 0.0040 & 0.0000 & 0.0548 & 0.2505 & 0.5816 & 0.0000 & 0.8230 & 0.9694 & 0.2960 & 0.0000 & 0.6190 \\
 & Runtime (s) & 4.0745 & 4.1487 & 2.8978 & 2.8657 & 2.8689 & 2.7789 & 2.8218 & 2.8655 & 2.9227 & 2.8502 & 2.7969 & 4.1867 & 4.1376 \\
\midrule
DAGMM & F1 & 0.1373 & 0.1373 & 0.0510 & 0.0797 & 0.1056 & 0.2847 & 0.4614 & 0.0000 & 0.5354 & 0.6916 & 0.3280 & 0.0005 & 0.5602 \\
 & FPR & 0.0003 & 0.0003 & 0.0003 & 0.0003 & 0.0003 & 0.0003 & 0.0003 & 0.0003 & 0.0003 & 0.0003 & 0.0003 & 0.0003 & 0.0003 \\
 & Precision & 0.4000 & 0.4000 & 0.2889 & 0.3711 & 0.4725 & 0.7065 & 0.8932 & 0.0000 & 0.8049 & 0.9447 & 0.5887 & 0.0148 & 0.8354 \\
 & Recall & 0.0995 & 0.0995 & 0.0297 & 0.0484 & 0.0647 & 0.1950 & 0.3405 & 0.0000 & 0.4402 & 0.5915 & 0.2575 & 0.0002 & 0.4650 \\
 & Runtime (s) & 1.2303 & 1.2598 & 1.0350 & 1.0380 & 1.0437 & 1.0165 & 1.0159 & 1.0391 & 1.0477 & 1.0355 & 1.0286 & 1.2662 & 1.2555 \\
\midrule
DeepAnt & F1 & 0.0459 & 0.0459 & 0.1382 & 0.0389 & 0.1115 & 0.3916 & 0.1349 & 0.0250 & 0.1853 & 0.5744 & 0.0475 & 0.0130 & 0.2974 \\
 & FPR & 0.0001 & 0.0001 & 0.0000 & 0.0000 & 0.0000 & 0.0004 & 0.0001 & 0.0000 & 0.0002 & 0.0008 & 0.0001 & 0.0000 & 0.0003 \\
 & Precision & 0.1104 & 0.1104 & 0.2000 & 0.0500 & 0.2667 & 0.9916 & 0.1826 & 0.0500 & 0.2811 & 0.7242 & 0.0493 & 0.0444 & 0.3632 \\
 & Recall & 0.0389 & 0.0389 & 0.1200 & 0.0347 & 0.0814 & 0.2660 & 0.1213 & 0.0167 & 0.1631 & 0.5204 & 0.0462 & 0.0081 & 0.2808 \\
 & Runtime (s) & 2.5337 & 2.5726 & 2.0042 & 2.0030 & 1.9948 & 1.9854 & 1.9883 & 2.0092 & 2.0375 & 1.9903 & 1.9789 & 2.6261 & 2.5564 \\
\midrule
IForest & F1 & 0.0412 & 0.0412 & 0.0000 & 0.0051 & 0.0000 & 0.0000 & 0.0493 & 0.0000 & 0.2377 & 0.2134 & 0.1271 & 0.0000 & 0.0730 \\
 & FPR & 0.0000 & 0.0000 & 0.0000 & 0.0000 & 0.0000 & 0.0000 & 0.0000 & 0.0000 & 0.0000 & 0.0000 & 0.0000 & 0.0000 & 0.0000 \\
 & Precision & 0.1444 & 0.1444 & 0.0000 & 0.0167 & 0.0000 & 0.0000 & 0.1500 & 0.0000 & 0.4333 & 0.4833 & 0.3333 & 0.0000 & 0.3000 \\
 & Recall & 0.0276 & 0.0276 & 0.0000 & 0.0030 & 0.0000 & 0.0000 & 0.0360 & 0.0000 & 0.1802 & 0.1531 & 0.0888 & 0.0000 & 0.0470 \\
 & Runtime (s) & 0.1261 & 0.1319 & 0.1198 & 0.1187 & 0.1199 & 0.1226 & 0.1183 & 0.1215 & 0.1204 & 0.1181 & 0.1290 & 0.1360 & 0.1291 \\
\midrule
TGAN-AD & F1 & 0.1166 & 0.1162 & 0.0970 & 0.0452 & 0.0680 & 0.2300 & 0.3736 & 0.0033 & 0.5479 & 0.5648 & 0.2837 & 0.0017 & 0.4683 \\
 & FPR & 0.0027 & 0.0027 & 0.0040 & 0.0042 & 0.0040 & 0.0040 & 0.0041 & 0.0041 & 0.0041 & 0.0041 & 0.0041 & 0.0027 & 0.0027 \\
 & Precision & 0.2146 & 0.2035 & 0.2229 & 0.2103 & 0.2894 & 0.5866 & 0.7223 & 0.0153 & 0.8440 & 0.8125 & 0.5122 & 0.0083 & 0.7010 \\
 & Recall & 0.0922 & 0.0921 & 0.0678 & 0.0281 & 0.0440 & 0.1567 & 0.2824 & 0.0019 & 0.4595 & 0.4800 & 0.2240 & 0.0010 & 0.3918 \\
 & Runtime (s) & 9.8253 & 10.0572 & 6.9986 & 6.9556 & 6.9946 & 6.8101 & 6.8623 & 7.0094 & 7.0734 & 6.9263 & 6.8951 & 10.2985 & 10.2137 \\
\bottomrule
\end{tabular}
}
\end{table}
    \clearpage
}

\begin{table*}[t]
\centering
\caption{Average False Positive Rate (FPR) on Clean Data by Data Generating Process (DGP). Models Sorted by FPR (Low to High).}
\label{tab:clean_fpr_sorted}
\scalebox{0.6}{
\begin{tabular}{lccccccccccc}
\toprule
DGP & AlioghliOkay & TimeGPT & DAGMM & TGAN-AD & GARCH & IForest & ReGEN-TAD & TranAD & DeepANT & OLS & RRR \\
\midrule
Static Factor Model & 0.0122 & 0.0361 & 0.0607 & 0.0629 & 0.0545 & 0.0921 & 0.0540 & 0.0828 & 0.0429 & 0.0989 & 0.1088 \\
Factor-GARCH & 0.0056 & 0.0366 & 0.0596 & 0.0544 & 0.0558 & 0.0530 & 0.0538 & 0.0683 & 0.0379 & 0.1039 & 0.1010 \\
GARCH & 0.0000 & 0.0361 & 0.0272 & 0.0418 & 0.0509 & 0.0073 & 0.0528 & 0.0186 & 0.0574 & 0.0977 & 0.1028 \\
IID Gaussian & 0.0037 & 0.0382 & 0.0485 & 0.0579 & 0.0530 & 0.0781 & 0.0526 & 0.0684 & 0.0297 & 0.0877 & 0.0894 \\
IID Student-t & 0.0168 & 0.0349 & 0.0636 & 0.0604 & 0.0515 & 0.0597 & 0.0577 & 0.0677 & 0.0286 & 0.0914 & 0.0974 \\
Volatility Drift & 0.0000 & 0.0503 & 0.0000 & 0.0001 & 0.0418 & 0.0000 & 0.0527 & 0.0000 & 0.4279 & 0.2638 & 0.2774 \\
VAR & 0.0073 & 0.0356 & 0.0577 & 0.0590 & 0.0515 & 0.0835 & 0.0538 & 0.0771 & 0.0339 & 0.0826 & 0.0754 \\
\midrule
Overall Average FPR & 0.0065 & 0.0382 & 0.0453 & 0.0481 & 0.0513 & 0.0534 & 0.0539 & 0.0547 & 0.0940 & 0.1180 & 0.1217 \\
\bottomrule
\end{tabular}
}
\end{table*}

\subsection{False Alarm Control Under In-Control Regimes}

We evaluate false-alarm behavior under strictly clean conditions ($\gamma = 0$), where no structural anomalies are injected. In this setting, any flagged observation is by construction a false positive, and performance is summarized by the False Positive Rate (FPR). This experiment isolates calibration stability and in-control robustness, ensuring that detection gains observed in contaminated settings are not offset by excessive alerts during economically stable regimes. All procedures are calibrated using a common nominal tail probability $\alpha = 0.05$ to ensure comparability of in-control false-alarm rates.

Clean panels are generated with cross-sectional dimension $p=100$ and length $T=500$ under seven economically interpretable data-generating processes (DGPs). These include: IID Gaussian and IID Student-$t$ innovations (light- and heavy-tailed homoskedastic benchmarks); a univariate GARCH(1,1) specification applied cross-sectionally to model volatility clustering; a Static Factor model with latent common components and fixed loadings; a Factor-GARCH design combining common-factor dependence with time-varying idiosyncratic volatility; a Stable VAR(1) process with coefficient matrix satisfying spectral radius constraints to ensure stationarity; and a Smooth Volatility Drift regime in which unconditional variance evolves gradually over time without discrete structural breaks. Parameter magnitudes are chosen to reflect realistic equity-return dispersion and persistence patterns rather than extreme stylization. Performance metrics are computed from 700 independent Monte Carlo replications spanning the seven DGP environments. 

Table~\ref{tab:clean_fpr_sorted} reports average FPR across DGPs, sorted from lowest to highest overall false-alarm rate. Several patterns emerge. AlioghliOkay exhibits the most conservative calibration (0.65\%), followed by TimeGPT (3.82\%). DAGMM, TGAN-AD, GARCH, IForest, ReGEN-TAD, and TranAD cluster tightly in the 4--6\% range, indicating broadly comparable baseline discipline. ReGEN-TAD maintains a stable average FPR of 5.39\% across heterogeneous environments, reflecting the stabilizing effect of rank-based calibration. 

By contrast, residual-based linear monitors display materially elevated alert rates: OLS averages 11.8\% and RRR 12.17\%. The primary source of instability is smooth volatility evolution rather than heavy tails. Under the Volatility Drift DGP, DeepANT (42.79\%), OLS (26.38\%), and RRR (27.74\%) exhibit pronounced false-alarm inflation, indicating difficulty distinguishing gradual second-moment evolution from structural abnormality. Heavy-tailed IID noise and stationary GARCH dynamics generate only modest FPR increases for most procedures.

Overall, false alarms are driven primarily by sensitivity to slowly evolving variance rather than excess kurtosis or cross-sectional dependence per se. From a financial monitoring perspective, evaluation under $\gamma = 0$ is indispensable: detection power must be interpreted jointly with in-control false-alarm discipline, since over-signaling during stable regimes can impose substantial economic costs in deployment.

\subsection{Factor Attribution in Sector Shock Experiments}
\label{sec:sector_attribution_sim}

To evaluate the interpretability of the proposed framework in a realistic financial setting, we conduct controlled sector-shock experiments using actual NASDAQ-100 return data with injected perturbations. Rather than simulating artificial cross-sectional structure, we preserve the empirical dependence, volatility clustering, and sectoral correlation present in observed equity returns and superimpose structured anomalies within a designated economic sector. This design ensures that attribution recovery is tested under realistic market geometry rather than stylized covariance constructions.

We construct a high-dimensional equity return panel using constituents of the NASDAQ-100 index, yielding $p=101$ assets spanning multiple sectors. Let $\mathbf{X}_t \in \mathbb{R}^p$ denote observed asset log-returns at time $t$. Overlapping rolling windows of length $L=36$,
\[
\mathbf{X}^{(L)}_t = \{\mathbf{X}_{t-L+1}, \ldots, \mathbf{X}_t\},
\]
serve as inputs to the anomaly detection model.

Sector membership is treated as known structural information. In each experiment, a single target sector $S \subset \{1,\dots,p\}$ is selected, and anomalies are injected exclusively into assets belonging to that sector over a fixed event window, while the remaining assets evolve according to their realized historical dynamics. We consider four economically motivated perturbation mechanisms: volatility spikes, mean shifts, persistent trends, and jump bursts. These perturbations are calibrated relative to the empirical return scale of the affected sector to avoid unrealistic magnitude distortions.

For each rolling window, ReGEN-TAD produces a binary anomaly indicator $\widehat{A}_t \in \{0,1\}$ and a vector of asset-level attribution scores
\[
\mathbf{a}_t = (a_{t,1}, \ldots, a_{t,p}),
\]
where larger values indicate greater contribution to the detected anomaly. For windows flagged as anomalous ($\widehat{A}_t = 1$), assets are ranked in descending order of attribution magnitude.

To quantify structured recovery, we define the \emph{sector match ratio} for window $t$ as
\[
\mathrm{MR}_t
=
\frac{1}{|S|}
\sum_{j \in S}
\mathbb{I}\!\left(
j \in \mathrm{Top}_{|S|}(\mathbf{a}_t)
\right),
\]
where $\mathrm{Top}_{|S|}(\mathbf{a}_t)$ denotes the indices of the $|S|$ largest attribution scores and $\mathbb{I}(\cdot)$ is the indicator function. This metric measures the proportion of truly perturbed sector assets recovered among the top-ranked contributors identified by the model. Let $\mathcal{A}$ denote the set of windows flagged as anomalous. Overall attribution recovery is summarized by the average match ratio
\[
\overline{\mathrm{MR}}
=
\frac{1}{|\mathcal{A}|}
\sum_{t \in \mathcal{A}} \mathrm{MR}_t.
\]

Table~\ref{tab:sector_match_ratio_small} reports sector-level attribution recovery across perturbation types. Under sustained or diffuse disturbances, recovery is strong. Volatility spikes in Information Technology yield an average match ratio of 0.693 across 38 sector assets, indicating that a substantial proportion of affected assets are correctly ranked among the top contributors. Mean shifts in Communication Services (sector size 10) produce an average match ratio of 0.574, while persistent trend perturbations in Consumer Discretionary (sector size 18) yield 0.551, demonstrating consistent sector localization.

In contrast, jump-burst anomalies within the Health Care sector (12 assets) produce a markedly lower average match ratio of 0.110. This attenuation reflects the inherent difficulty of attributing highly transient and localized shocks whose cross-sectional footprint is brief relative to the rolling window length and embedded within empirical market volatility.

Overall, the results demonstrate that the attribution mechanism reliably recovers economically meaningful sector structure when perturbations are sustained or propagate gradually through the panel. While recovery weakens for short-lived jump-type disturbances, the framework consistently localizes structured cross-sectional shocks within realistic financial data, supporting the interpretability of the proposed approach in high-dimensional equity panels.

\begin{table}[t]
\centering
\footnotesize
\setlength{\tabcolsep}{4pt}
\caption{Sector-level attribution recovery under structured perturbations. 
Anomalies are injected exclusively within a target sector, and recovery is measured 
via the match-ratio metric, defined as the proportion of affected assets 
identified among the top-ranked attribution scores.}
\label{tab:sector_match_ratio_small}
\begin{tabular}{l l c c}
\toprule
Sector & Anomaly Type & Avg.\ Match & Sector Size \\
\midrule
Information Technology & Volatility spike & 0.693 & 38 \\
Communication Services & Mean shift       & 0.574 & 10 \\
Consumer Discretionary & Trend            & 0.551 & 18 \\
Health Care            & Jump bursts      & 0.110 & 12 \\
\bottomrule
\end{tabular}
\end{table}

\subsection{Historical Market Episodes}

To assess external validity beyond controlled perturbations,
we evaluate ReGEN-TAD on unmodified Dow 30 returns during two
well-documented systemic crises: the 2008 global financial crisis
and the February--April 2020 COVID-19 market collapse.
No anomaly injection, recalibration, or threshold adjustment is performed.

\begin{figure*}[t]
    \centering
    \includegraphics[width=\textwidth]{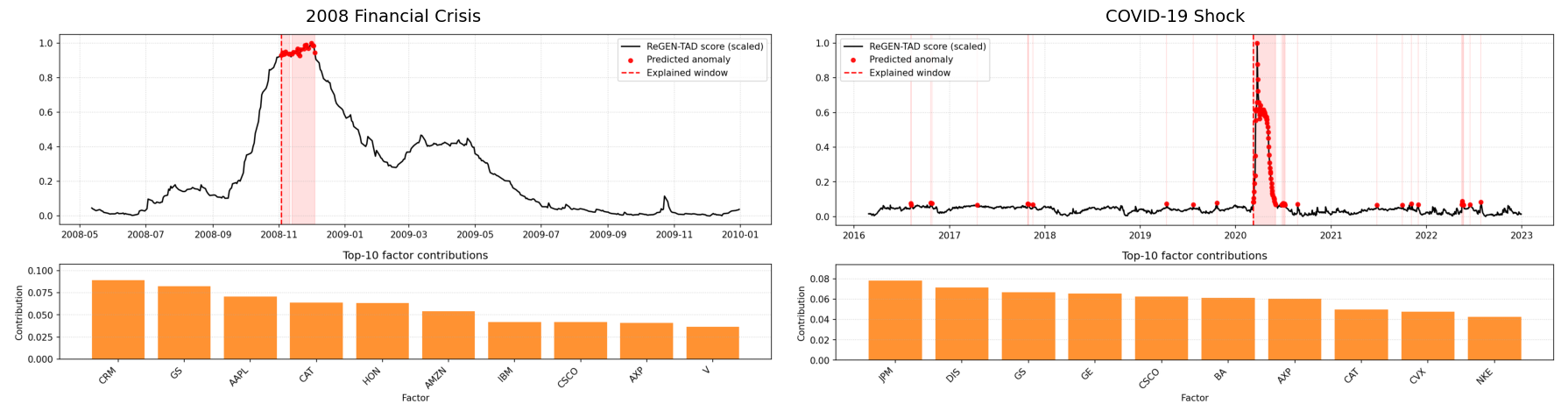}
    \caption{
    ReGEN-TAD anomaly score and attribution analysis during two
    historical systemic crises. Left: 2008 financial crisis.
    Right: COVID-19 market collapse.
    In both episodes, anomaly scores remain stable during
    pre-crisis periods and exhibit sharp concentration during
    systemic stress intervals. Attribution panels highlight
    economically coherent cross-sectional contributors consistent
    with broad market contagion.
    }
    \label{fig:historical_episodes}
\end{figure*}

Figure~\ref{fig:historical_episodes} displays the scaled anomaly
score and corresponding attribution analysis for both episodes.
In the 2008 case (left panel), the score remains low and stable
throughout 2007 and early 2008, followed by a sharp and concentrated
elevation beginning in September 2008, coinciding with the Lehman
bankruptcy and peak systemic stress. The score remains elevated
into early 2009 before gradually reverting toward baseline,
consistent with prolonged liquidity disruption.

The COVID-19 episode (right panel) exhibits a similarly abrupt
regime break in late February 2020. Anomaly detections cluster
within the collapse window, while pre- and post-crisis intervals
display comparatively low baseline activity. This concentration
indicates sensitivity to genuine structural dislocation rather
than routine volatility fluctuations.

The attribution panels further demonstrate economically coherent
cross-sectional structure. During 2008, contributions are broadly
distributed across cyclical, industrial, and financial constituents,
reflecting economy-wide contagion. During the COVID shock, financial
and consumer-exposed firms dominate the ranking, consistent with
documented stress transmission channels. In both cases, the framework
localizes structured exposure patterns rather than dispersing influence
uniformly across assets.

Collectively, these results indicate that ReGEN-TAD identifies
historically documented systemic crises without synthetic perturbations
and produces interpretable sector-level localization during real
market dislocations.

\section{Conclusion}
\label{sec:conclusion}

This paper develops ReGEN-TAD, an interpretable ensemble-based generative framework for anomaly detection in high-dimensional financial time series. The central contribution is not merely improved detection accuracy, but the integration of detection, calibration, and cross-sectional attribution within a single econometrically coherent architecture. The framework combines generative reconstruction signals and structured cross-sectional scoring into a unified ensemble, while enforcing a leakage-free rolling calibration protocol to ensure strictly out-of-sample evaluation.

Methodologically, the paper advances anomaly detection along three dimensions. First, it introduces a generative ensemble design that stabilizes detection performance across heterogeneous structural perturbations, including regime shifts, contagion dynamics, volatility clustering, and sector-localized shocks. Second, it embeds an explicit attribution mechanism that produces asset-level contribution scores aligned with economically meaningful sector and factor structure. This eliminates the need for ex post explanatory regressions and allows structural localization to emerge directly from the detection model. Third, the adoption of strict rolling calibration, with training, validation, and testing partitions using fixed thresholds and normalization based exclusively on past data, ensures financial-econometric credibility and guards against forward-looking contamination.

Empirical evidence across synthetic structural simulations, financially structured market regimes, and clean-regime specificity tests demonstrates that ReGEN-TAD achieves a rare combination of high recall and extremely low false positive rates. The model remains stable under diverse data-generating processes and avoids the false-alarm inflation observed in procedures that lack robust normalization. In sector shock experiments based on real equity panels with controlled perturbations, the attribution mechanism reliably recovers the affected cross-sectional structure under sustained disturbances, providing concrete evidence that interpretability is structurally grounded rather than heuristic.

Computationally, ReGEN-TAD is more intensive than the other benchmark procedures due to its multi-component generative backbone and ensemble aggregation. Runtime comparisons indicate that the method ranks behind simpler residual-based or single-architecture detectors in raw execution speed. This reflects a deliberate design trade-off: the framework prioritizes calibration stability, robustness to heterogeneous structural mechanisms, and structured attribution over minimal computational cost. In practical deployment, the added overhead remains tractable on modern hardware and is incurred only during model training, while inference operates at standard neural-network speeds.

Taken together, the results suggest that anomaly detection in high-dimensional financial panels should not be treated as a purely predictive exercise. Detection, calibration discipline, and structural attribution must be designed jointly. By embedding economically meaningful localization within a rigorously calibrated generative ensemble, ReGEN-TAD offers a scalable and interpretable alternative to black-box detection systems and residual-only econometric monitors. The framework therefore bridges modern generative modeling with classical financial econometric principles, advancing both detection reliability and structural transparency in complex market environments.

\newpage

\section*{Disclosure of Interest}

The authors declare that they have no competing interests.

\section*{Data Availability Statement}

All data generation scripts, simulation code, and implementation details necessary to reproduce the results in this paper are publicly available at:
\url{https://github.com/martinwg/ReGEN-TAD}.

\section*{Funding}

No funding was received for this research.


\bibliographystyle{unsrtnat}
\bibliography{references}  

\end{document}